\documentclass[journal]{IEEEtran}
\ifCLASSINFOpdf
  \usepackage[pdftex]{graphicx}
\else
\fi
\usepackage{blindtext}

\usepackage[cmex10]{amsmath}
\usepackage{algorithm}
\usepackage{algorithmic}

\usepackage{amsfonts}
\usepackage{multirow}
\usepackage{comment}
\usepackage{hyperref}

\usepackage{xcolor}
\usepackage{bm}
\usepackage{booktabs}
\usepackage{xfrac}
\makeatletter

\usepackage{tikz}
\usepackage{textcomp}
\usepackage{hyperref}
\usepackage{lipsum}

\newcommand\copyrighttext{%
  \footnotesize \textcopyright 2021 IEEE. Personal use of this material is permitted.
  Permission from IEEE must be obtained for all other uses, in any current or future
  media, including reprinting/republishing this material for advertising or promotional
  purposes, creating new collective works, for resale or redistribution to servers or
  lists, or reuse of any copyrighted component of this work in other works.}
\newcommand\copyrightnotice{%
\begin{tikzpicture}[remember picture,overlay]
\node[anchor=south,yshift=0pt] at (current page.south) {\fbox{\parbox{\dimexpr\textwidth-\fboxsep-\fboxrule\relax}{\copyrighttext}}};
\end{tikzpicture}%
}

\newcommand\fs@norules{\def\@fs@cfont{\bfseries}\let\@fs@capt\floatc@ruled
  \def\@fs@pre{}%
  \def\@fs@post{}%
  \def\@fs@mid{\kern3pt}%
  \let\@fs@iftopcapt\iftrue}
\makeatother
\floatstyle{norules}
\restylefloat{algorithm}

\usepackage{makecell}
\begin{document}
%
\title{PAMELI: A Meta-Algorithm for Computationally Expensive Multi-Objective Optimization Problems}
%
%
%

\author{ Santiago Cuervo,~Miguel Melgarejo,~\IEEEmembership{Senior Member,  IEEE},
        Angie Blanco-Ca\~non,
        Laura Reyes-Fajardo~and
        Sergio Rojas-Galeano}
\maketitle
\copyrightnotice
\begin{abstract}
We present an algorithm for multi-objective optimization of computationally expensive problems. The proposed algorithm is based on solving a set of surrogate problems defined by models of the real one, so that only solutions estimated to be approximately Pareto-optimal are evaluated using the real expensive functions. Aside of the search for solutions, our algorithm also performs a meta-search for optimal surrogate models and navigation strategies for the optimization landscape, therefore adapting the search strategy for solutions to the problem as new information about it is obtained. The competitiveness of our approach is demonstrated by an experimental comparison with one state-of-the-art surrogate-assisted evolutionary algorithm on a set of benchmark problems.
\end{abstract}

\begin{IEEEkeywords}
Multi-objective Optimization, Evolutionary algorithms, Surrogate Assisted Optimization, Meta-optimization, Landscape Analysis.
\end{IEEEkeywords}

%
\IEEEpeerreviewmaketitle

\section{Introduction}
%
%
%
%

\IEEEPARstart{O}{ptimization} problems where there are not closed-form expression for fitness functions suppose a hard task for conventional heuristics. Surrogate models have been proposed to deal with this kind of functions. However, selecting an appropriate model for a particular function is not trivial because of lack of knowledge about the problem or computational restrictions around it. Besides, fitness landscapes are not often homogenous which poses an additional difficulty since no general surrogate model can be considered for all objective functions. Although different surrogate models for these functions might be inferred, the question about how to effectively navigate their corresponding optimization landscapes is not solved in consequence.

Many optimization problems require the simultaneous optimization of multiple objectives, as there are usually several criteria under which a solution should be evaluated. These kind of problems are known as \textit{Multi-objective Optimization Problems} (MOPs). Some of the objectives can be conflicting in nature, and therefore, typically a single optimal solution does not exist, but multiple so-called \textit{Pareto-optimal} solutions, which are the best trade-off solutions between objectives. The set of all Pareto-optimal solutions in the search space is called the \textit{Pareto Set} (PS) and its image in the objective space is called the \textit{Pareto Front} (PF).

In addition, the difficulty of choosing a single algorithm that solves efficiently learning and optimization problems is a natural consequence of the No-Free-Lunch Theorems (NFLTs) \cite{Wolpert96a} \cite{Wolpert97a}, which state that an algorithm performs better than random search only in some class of problems, and therefore, there is no theoretical basis to choose one without a priori knowledge of the problem where there is to be applied. An extension of these theorems for MOPs was proposed in \cite{service2010no}, where they deal with  optimization functions that are structurally \textit{closed under permutation}, a condition that is guaranteed with Pareto optimality. What these theorems imply is that selecting an appropriate algorithm requires acquiring prior knowledge that allows to make assumptions about the problem at hand.

Given that in MOPs the desired result is the PS,
evolutionary algorithms have arisen as a suitable approach to find this set, because they are able to evolve a population of solutions in a single execution. 
Research in their application to MOPs, termed \textit{Evolutionary Multi-Objective Optimization} (EMO), has been flourishing in academia for over two decades \cite{ZHOU201132}.
However, problems which need to be solved with a limited budget of function evaluations pose significant challenges to conventional EMO algorithms. A closed-form expression for the objective functions may not be known, and instead, computationally expensive numerical simulations or costly physical experiments must be performed to evaluate the population of solutions \cite{Chugh17a}.

In this regard, one popular approach to address computationally expensive problems in EMO are \textit{Surrogate-Assisted Evolutionary Algorithms} (SAEAs), where regression models are trained to replace the objective functions at some points of the evaluation \cite{JIN201161}, with the advantage of being more computationally efficient.  These models, known as \textit{surrogates}, can be any type of regression methods, but, as several studies have shown \cite{Goel2007}\cite{Samad2008}\cite{Montemayor11a}, their choice heavily influences SAEAs performance. Unfortunately, there is no universal criteria for choosing a regression model that suitably approximates the optimization landscape \cite{krvea}. 

Regarding navigation of the optimization landscape, since there is no single best optimization algorithm for all problems, the effectiveness of one algorithm to produce solutions would depend on the properties of such landscape \cite{Pham13a} \cite{Jones1995} (e.g. convexity, separability, multi-modality, etc.). An ill-suited optimization algorithm may result in slow convergence or even not convergence at all to a good enough result, that is, a solution set far away from the true PS in the case of MOP.

In this context, we propose a meta-algorithm that goes beyond the framework of SAEAs. It starts with some weak priors, encoded in a heterogeneous set of models coupled with optimization algorithms, and refines them while more information about the problem landscape properties is discovered as a byproduct of the search for solutions. This is carried out not only by tuning an adequate surrogate model for each objective function but also selecting the companion search algorithm that maximizes the fitness of the solution set. Our aim is, in accordance with the NFLTs, to not strive for generalization but rather for specialization, by allowing the meta-algorithm to adapt to the specifics of the problem. In this sense we hope to find better suited combinations of surrogate models and optimization algorithms that could lead to faster convergence on computationally expensive optimization problems. 

The rest of this paper is organized as follows: section \ref{sec:background} introduces some concepts related to the work at hand that seek to contextualize the reader on the problem and proposed solution. Next, a detailed description of the proposed algorithm is provided in section \ref{sec:algorithm}. Numerical results of the proposed meta-algorithm on a set of benchmark problems and a comparison with a state-of-the-art SAEA are presented in section \ref{sec:num_ex} and discussed in section \ref{sec:discussion}. Finally, in section \ref{sec:conclusions} we present our conclusions and suggest future work directions.

\section{Background}
\label{sec:background}

\subsection{Surrogate Models}

In many engineering and science problems, it is necessary to use computer simulations to model physical phenomena incurring in a fraction of the cost to perform real experiments \cite{Diaz-Manriquez2016}. The computational cost of a simulation increases with the complexity of the model \cite{Goh2011}. In a similar manner, in many optimization problems where objective functions are computationally expensive, the use of surrogate models has been proposed to replace them during the evaluation of some candidate solutions \cite{JIN201161}.

A surrogate is a regression model that approximates the objective function using partial  information obtained before and during the process of solving the problem \cite{Bhattacharjee2016}. Surrogate models are intended to be computationally simpler than the actual objective functions \cite{Ong2005}, because they capture a possible relation between input and output variables, but not the underlying process \cite{Diaz-Manriquez2016}. 

\subsection{Surrogate Assisted Evolutionary Algorithms}
\label{sec:background_saeas}

SAEAs rise as the combination of surrogate models with evolutionary algorithms. Many surrogate models have been used with SAEAs, where the most popular are Kriging based approaches \cite{Forrester2008}, Artificial Neural Networks (ANNs) \cite{Kourakos2013}, Support Vector Machines (SVMs) \cite{Herrera2014}, Radial Basis Functions (RBFs) \cite{Habib2019} and Polynomial Response Surface Methods (RSMs) \cite{Liu2008}. Aside of the choice of the surrogate model, a major challenge in SAEAs is \textit{model-management}, that it refers to the strategy to decide which solutions should be evaluated using the objective function or the surrogate model. Besides, model-management also deals with the update of the surrogate models  \cite{JIN201161}.

Several SAEAs stand out for multi-objective optimization. ParEGO \cite{Knowles2006} is an extension for MOPs of the \textit{Efficient Global Optimization} (EGO) \cite{Jones1998} algorithm, which uses Kriging to model the optimization landscape from solutions visited during the search for Pareto-optimal solutions. The \textit{Kriging - Reference Vector guided Evolutionary Algorithm} (K-RVEA) \cite{krvea} also uses Kriging models as surrogates in conjunction with a model-management based on reference vectors. K-RVEA was shown to outperform ParEGO in several benchmark problems. 

Recently, in \textit{Hybrid Surrogate-assisted Many-objective Evolutionary Algorithm} (HSMEA) \cite{Habib2019} the issue of the choice of the surrogate model was considered. HSMEA approximates objective functions using Kriging, RSM and RBFs, and selects among them the one with lower approximation error to guide the search at each iteration. Experimental results showed that HSMEA exhibited superior performance than K-RVEA and other state-of-the-art SAEAs.

\subsection{Heterogeneity in optimization landscapes}
The \textit{optimization landscape} is a metaphor frequently used to describe the manifold obtained from mapping the search space to the objective space in optimization problems. 
The properties of the optimization problem such as convexity, continuity or, multi-modality, 
produce a diversity of landscapes,  suggesting the need of considering heterogeneous optimization algorithms in order to solve them appropriately.

\begin{figure}[!ht]
\centering
\begin{tabular}{cc}
  \includegraphics[width=0.23\textwidth]{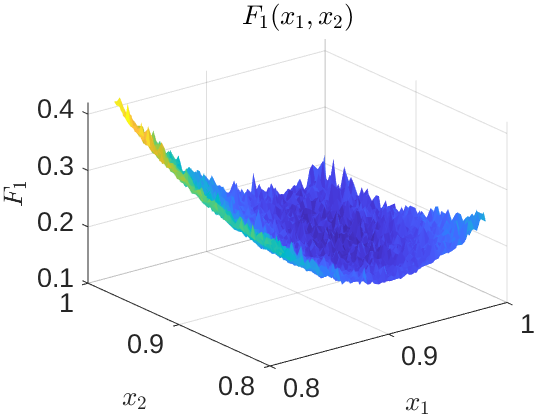}   & \includegraphics[width=0.23\textwidth]{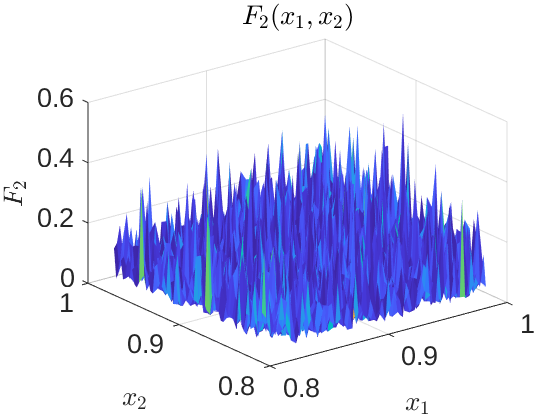} \\
  (a)   & (b)
\end{tabular}

\includegraphics[width=0.23\textwidth]{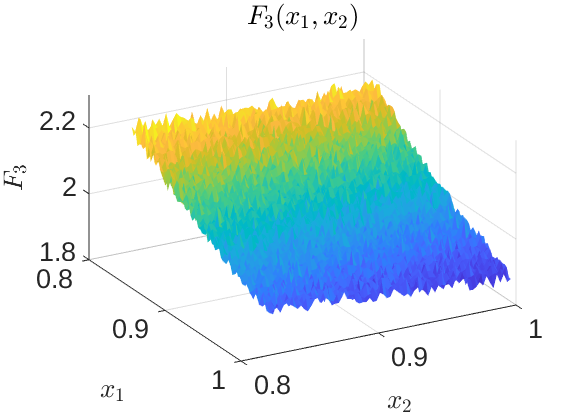}\\
(c)
\caption{Optimization landscapes of the three objective functions of the DDMOP3 problem of the \textit{CEC Online Data-Driven Multi-Objective Optimization Competition 2019} \cite{cec0319} with respect to the first two decision variables.}
\label{fig:landscapes}
\end{figure}

As an example let us consider a MOP for the design of a switching ripple suppressor with three objective functions \cite{cec0319}. We illustrate the optimization landscapes obtained from the first two decision variables in Fig. \ref{fig:landscapes}. It can be seen that the landscapes depicted in Fig. \ref{fig:landscapes}a and Fig. \ref{fig:landscapes}c are roughly convex and planar, respectively, hence a gradient-based strategy performing narrow exploration, like \textit{stochastic gradient-descent}, would be a good optimizer. On the other hand, the landscape depicted in Fig. \ref{fig:landscapes}b is noisier and  plagued of local minima, and an algorithm performing a wider exploration, such as a population-based technique, would be more convenient. 

Regarding surrogate modeling, the properties of the optimization landscape should also be taken into account. Whereas  simple quadratic and linear models would suffice for the functions in Fig. \ref{fig:landscapes}a and Fig. \ref{fig:landscapes}c, the function depicted in Fig. \ref{fig:landscapes}b requires a more complex model, like a large neural network, to approximate its highly multimodal structure. Thus, the common practice of using only one type of surrogate model in SAEAs \cite{Jones1998}\cite{krvea} disregards the heterogeneity of landscapes that can appear in a  MOP.

Consequently, our main motivation is to address the issue of choosing the best suited combination of optimizer and surrogate model in problems in which there is not enough a priori knowledge about the properties of the optimization landscape. As it will be explained in the next section, our philosophy is to face this issue with a strategy that ranks pairs of surrogate and optimizer according to their performance during the interaction with the optimization problem.

\subsection{Meta-optimization and Hyper-heuristics}
Parameter tuning is usually a necessary step when applying metaheuristics, and it is often a process that is carried out manually, by trial and error, and/or using expert knowledge. \textit{Meta-optimization} refers to the use of a metaheuristic to tune the parameters of another metaheuristic \cite{Neumuller12a}. It has been used in both, single-objective optimization problems \cite{Neumuller12a} \cite{Crawford17a}, and MOPs \cite{Ugolotti19a} \cite{Lapa19a}. 

A related concept within the literature is \textit{Hyper-heuristics} \cite{Burke13a}, which is a further extension to the framework of meta-optimization. In this concept, a metaheuristics is used to search for algorithms rather than parameters. Different heuristics or heuristic components can be selected, generated or combined to solve single-objective and multi-objective optimization problems \cite{Maashi15a}. Our search for the best suited combination of optimizer and surrogate model could be considered an Hyper-heuristic that searches in the joint space of multi-objective metaheuristics and regression methods.

\begin{figure*}[!t]
\centering
\includegraphics[width=0.9\textwidth]{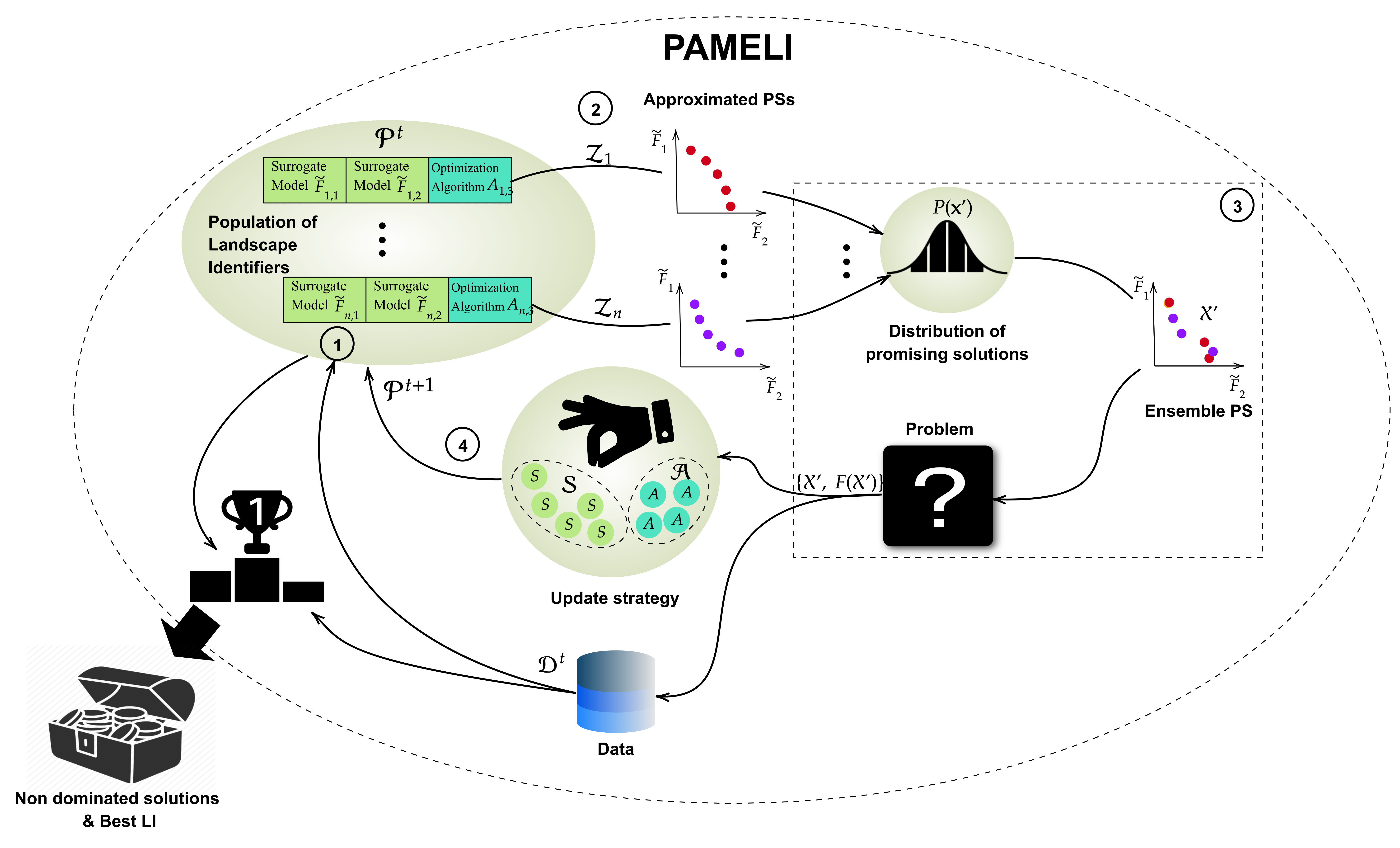}
\caption{Schematic of one iteration of the PAMELI algorithm on a problem with $m=2$ objectives and a population $\mathcal{P}^t$ of $n$ LIs. After initialization, an iteration proceeds as follows: 1. The surrogate models ($\tilde{F}_{i, j}$) of each LI are trained using the dataset $\mathcal{D}^t$ 2. The optimization algorithms ($A_{i, m+1}$) are used to obtain $n$ approximations to the PS (APSs) 3. A probabilistic model of promising solutions is build using the APSs and is evaluated in the actual objective space 4. The information obtained is added to the dataset and used to select a new population of LIs from the sets $\mathcal{S}$ and $\mathcal{A}$. The output of PAMELI consists of all non dominated solutions in the dataset $\mathcal{D}^t$ and the best LI in $\mathcal{P}^t$.
}
\label{fig:li_diagram}
\end{figure*}


\section{Proposed algorithm}
\label{sec:algorithm}
\subsection{General description}

The motivation of our proposal is twofold. On the one hand, from a theoretical point of view, we intend to consider
the consequences of the NFLTs regarding the convenience of adapting optimization algorithms to the specifics of a given problem, rather than attempting to succeed in a broader class of problems. On the other hand, from a pragmatic perspective, we aim to reduce the number of objective function evaluations needed for convergence to the PF on computationally expensive MOPs.

For this purpose, we introduce the \emph{landscape identification} approach as a way to cope with an MOP. It is inspired in the nonlinear dynamic system identification practice, where the inputs of the system are excited in order to sample its outputs so that a model of the underlying dynamics can be updated \cite{Wang1995}. The structure of the model is tuned as new information is collected from sampling providing a better insight of the actual system. In some cases, sampling can be guided as the  model is improved allowing to explore new regions of the dynamics \cite{Schoukens2019}. Likewise, identification of  optimization landscapes can be understood in a similar manner. It considers a surrogate model of the  landscape that is refined as more information is taken from the actual problem. This refined model is sampled by means of an optimization algorithm that explores promising regions to acquire useful information from reality. We call the combination of a model and an optimization algorithm a \emph{landscape identifier} (LI).

A LI produces as output an \emph{Approximated PS} (APS). The ensemble of multiple APSs obtained from a set of LIs is termed here as the \emph{Ensemble PS} (EPS). This ensemble is the set of solutions to be evaluated using the actual objective functions, therefore guiding the exploration towards promising regions of the optimization landscape.

A second exploratory process on the space of LIs can be carried out in order to look for the surrogate models and optimization algorithms that produce the best EPS. To this process we refer as \textit{meta-exploration}. The results of meta-exploration are the LIs to be used in future landscape identification processes.

We synthesize these ideas in our proposed algorithm for multi-objective optimization, called \emph{Pareto set Approximation by Meta-Exploration of Landscape Identifiers} (PAMELI). The pseudocode of PAMELI is presented in Algorithm \ref{algo:pameli} and Fig. \ref{fig:li_diagram} illustrates its components and the relations between them. Next, we give a detailed explanation of each component of the algorithm.

\begin{algorithm}[h]
 \caption{PAMELI.}
 \label{algo:pameli}
 \begin{algorithmic}[1]
    \STATE Initialize population of LIs and dataset.
    \FOR {number of iterations}
        \FOR {\textbf{each} LI}
            \STATE Train surrogate model.
            \STATE Identify PS. 
        \ENDFOR
        \STATE Build and evaluate the EPS.
        \STATE Update population of LIs.
    \ENDFOR
    \ENSURE Non dominated solutions in dataset and best LI.
 \end{algorithmic} 
 \end{algorithm}

\subsection{Multi-objective optimization at a glance}
A generic MOP is formulated as: 

\begin{equation}
    \min_{\mathbf{x}} \; F(\mathbf{x}) = \min_{\mathbf{x}} \; [F_{1}(\mathbf{x}), F_{2}(\mathbf{x}), ..., F_{m}(\mathbf{x})]^T, \; \forall \mathbf{x} \in \Omega,
    \label{eq:mop}
\end{equation} where $\Omega \subseteq \mathbb{R}^d$ is the $d$-dimensional \textit{search space}, and $\mathbf{x}$ is a \textit{solution}. $F(\mathbf{x})$ consists of $m$ objective functions, $F_i: \Omega \to \mathbb{R}, \; i=1,2,...,m$, so that, $F: \Omega \to \mathcal{F}$, with $\mathcal{F} \subseteq \mathbb{R}^{m}$ being the \textit{objective space}. Here, the goal of optimization is considered in the sense of \textit{Pareto optimality} \cite{ZHOU201132}, formally defined as follows:

\textbf{Definition 1.} A vector $\mathbf{u} = [u_1,u_2,...,u_m]$ is said to \textit{dominate} another vector $\mathbf{v} = [v_1,v_2,...,v_m]$, $\mathbf{u} \prec \mathbf{v}$, if $\forall i \in \{1,...,m\}, u_i \leq v_i$, and $\mathbf{u} \neq \mathbf{v}$.

\newcommand{\nexists}{\not\exists}
\textbf{Definition 2.} A solution $\mathbf{x} \in \Omega$ is \textit{Pareto optimal} if $\nexists \, \mathbf{y} \in \Omega$ such that $F(\mathbf{y}) \prec F(\mathbf{x})$. 

The set of all Pareto optimal solutions is called the \textit{Pareto Set} (PS) and its image in the objective space is called the \textit{Pareto Front} (PF). Solving a MOP consists then on finding the PS. 

In many MOPs is not possible to compute the exact PS \cite{hu2013a} and heuristic approaches that obtain estimations to it are used instead. Heuristic approaches compute a finite set of solutions that are assumed to be non-dominated but with no guarantee of their Pareto-optimality \cite{abounacer14a}. In addition, it is also desired for the set of solutions to be uniformly distributed on the PF, so as to represent a wider variety of trade-offs between objectives \cite{ZHOU201132}.

\subsection{LIs and data structures}

LIs correspond to the combination of two interrelated components: a \emph{surrogate model} and an \emph{optimization algorithm}. In the particular setting of multi-objective optimization, the surrogate model consists of a set of surrogate functions, each one approximating one of the objectives in the MOP; likewise, the optimization algorithm can be any population-based technique appropriate for MOPs. 
Furthermore, by means of an underlying genetic algorithm, PAMELI maintains and evolves a population of LIs, that fosters adaptation to the multi-objective optimization landscape. 

At a given iteration $t$, we define $\mathcal{X}^t = \{\mathbf{x}_1, \mathbf{x}_2, \ldots, \mathbf{x}_{\ell_t} \}$ as the incremental set of ${\ell_t}$ samples from the solution space, whose image in the objective space has been evaluated up to iteration $t$; thus, the working dataset at iteration $t$ would consist of this image plus the set of samples, i.e. $\mathcal{D}^t = \{\mathcal{X}^t, F(\mathcal{X}^t)\}$. The incremental property implies that for any two iterations $t$ and $t'$, with $t < t'$, then $\ell_{t'} > \ell_t$. 

Let us call the pool of surrogates  $\mathcal{S} = \{S_1, S_2, \ldots, S_{\ell_\mathcal{S}}\}$ as the set of $\ell_\mathcal{S}$ arbitrary regression methods that would be used by the LIs. We remark that the working set of regression methods on a given iteration $t$ will be fitted to $\mathcal{D}^t$, in order to obtain the actual surrogates of the  multi-objective functions, i.e. $\widetilde{F}_j(\mathbf{x})  \equiv S_i(\mathbf{x}  | \mathcal{D}^t)$, $j = 1,2,\ldots,m$. When clear from the context, we will refer to $\widetilde{F}_j$ as a given surrogate instance at iteration $t$.

On the other hand, we also define a pool of optimization algorithms $\mathcal{A} = \{A_1, A_2, \ldots, A_{\ell_\mathcal{A}}\}$ as an arbitrary set consisting of a number of $\ell_A$ MOP algorithms. Along these lines, the $i$-th landscape identifier would be defined as the tuple 

\begin{equation}
    LI_i = [\widetilde{F}_{i,1}, \widetilde{F}_{i,2}, \ldots, \widetilde{F}_{i,m}\, , A_{i, m+1}]
    \label{eq:li_structure}
\end{equation}

Therefore, a population of LIs at iteration $t$, $\mathcal{P}^t = \{LI_1, LI_2, \ldots, LI_n\}$ would be evolved by the meta-algorithm. In summary, PAMELI operates by simultaneous adaptation over the two data structures, $\mathcal{P}^t$ and $\mathcal{D}^t$. These data structures along with the adaptation procedure are depicted in Fig. \ref{fig:li_diagram}, tagged with labels 1 and 4, respectively. The following sub-sections explain in detail how such adaptation is accomplished.

\subsection{Initialization}
An initial population of LIs
of size $n$, $\mathcal{P}^0$, is generated. The genotypes of each LI are obtained by choosing $m$ elements from the set $\mathcal{S}$ and one element from the set $\mathcal{A}$ uniformly; these sets should be defined considering a-priori knowledge about the optimization landscape. In addition, a representative sample of the search space $\mathcal{X}^0$ is generated and is then evaluated with $F$ to obtain the initial dataset $\mathcal{D}^0$.

\subsection{Training of the surrogate models}

For the $i$-th LI, the training of $ \widetilde{F}_{i,j}: \Omega \to \mathbb{R}$ is performed in the usual manner of regression fitting by solving the following optimization problem:

\begin{equation}
    \min_{\Theta} L(\widetilde{F}_{i,j}(\mathbf{x}|\Theta), F_j(\mathbf{x})), \; \forall \mathbf{x} \in \mathcal{X}^t,
    \label{eq:trn_problem}
\end{equation} where $\Theta$ is the set of trainable parameters of the surrogate model and $L$ is its associated loss function. 

\subsection{Identification of the PSs}

The surrogate models of the $i$-th LI are used to formulate a \emph{surrogate multi-objective optimization problem}:

\begin{equation}
    \min_{\mathbf{x}} \; \widetilde{F}_i(\mathbf{x}) = \min_{\mathbf{x}} \; [\widetilde{F}_{i,1}(\mathbf{x}), \widetilde{F}_{i,2}(\mathbf{x}), ..., \widetilde{F}_{i,m}(\mathbf{x})]^T
    \label{eq:smop}
\end{equation}

The optimization algorithm of the LI, $A_{i, {m+1}}$, is used to approximately solve equation \ref{eq:smop}; as a result, $n$ APSs will be obtained.

\subsection{Building and evaluating the EPS}
\label{sec:sampling}
At each iteration $t$ the EPS is created as an ensemble of the $n$ APSs, consisting of a set of promising solutions $\mathcal{X}' = \{\mathbf{x}'_1, ..., \mathbf{x}'_{\ell'}\}$. Inspired by \emph{estimation of distribution algorithms} \cite{Hauschild11a}, we opted to build the EPS by sampling it from a mixture distribution $P(\mathbf{x}')$ estimated from the candidate solutions discovered by all LIs. We consider that this strategy is a convenient information sharing mechanism between the exploration processes that the LIs are conducting. 

Let us denote $\mathcal{Z}_i = \{\mathbf{z}_1, \ldots, \mathbf{z}_{\ell_i}\}$ as the set of candidate solutions in the APS of $LI_i$. Then, for $\mathcal{Z}_i$ we compute its $d$-dimensional sample mean vector, $\bm{\mu}_i$, as follows: 

\begin{equation}
    \bm{\mu}_i = [\mu_{i, 1}, \mu_{i, 2}, ..., \mu_{i, d}], \;
    \mu_{i, k} = \frac{1}{|\mathcal{Z}_i|} \sum_{\mathbf{z} \in \mathcal{Z}_i} \mathbf{z}_k
    \label{eq:sample_mean},
\end{equation} where $\mathbf{z}_{k}$ is the value of the $k$-th decision variable of the solution vector $\mathbf{z}$.

In addition, let $\mathcal{Z} = \bigcup\limits_{i=1}^{n} \mathcal{Z}_{i}
$, and its $d$-dimensional sample mean vector as:

\begin{equation}
 \bm{\mu} = [\mu_1, ..., \mu_d], \, \mu_{k} = \frac{1}{|\mathcal{Z}|} \sum_{\mathbf{z} \in \mathcal{Z}} \mathbf{z}_k, 
\end{equation}

We also compute a sample covariance matrix $\bm{\Sigma}$, where its $(k, l)$ entry is calculated as: 

\begin{equation}
    \sigma^2_{k, l} = \frac{1}{|\mathcal{Z}|} \sum_{\mathbf{z} \in \mathcal{Z}} (\mathbf{z}_k - \mu_{k})(\mathbf{z}_{l} - \mu_{l}),
\end{equation}

Lastly, the the set $\{\bm{\mu}_1, ... \bm{\mu}_n\}$ and $\bm{\Sigma}$ are used to sample the previously mentioned mixture distribution following the procedure described in Algorithm \ref{algo:sampling_eps}, where $\mathbf{x}'$ is a promising candidate solution, and $G(\cdot)$ is a  distribution model with mean $\bm{\mu}_i$ and covariance $\bm{\Sigma}$. The resulting $\mathcal{X}'$ is evaluated using the actual objective functions, obtaining its image in the objective space, $F(\mathcal{X}')$.

\begin{algorithm}[h]
 \caption{Build the EPS.}
 \label{algo:sampling_eps}
 \begin{algorithmic}[1]
    \FOR {$1, ..., \ell$}
        \STATE With uniform probability select a number $i$ from $\{1, .., n\}$
        \STATE Sample a solution from the distribution $G(\mathbf{x}' | \bm{\mu}_i, \bm{\Sigma})$ and append to $\mathcal{X}'$
    \ENDFOR
    \ENSURE $\mathcal{X}'$
 \end{algorithmic} 
 \end{algorithm}

\subsection{Update of the population of LIs}
The evaluation of the EPS in the objective space provides new information about the optimization landscape that can be used to update the population of LIs to better suit the problem. The update strategy can be based on a priori knowledge about the suitability of LIs conditioned on the information about the problem, or on the online performance of the population. In this study we opted for the latter, and the update is performed by means of an evolutionary algorithm. Other possible update strategies are discussed in section \ref{sec:discussion}.

We define the fitness function of the $i$-th LI, $E_i$, as:

\begin{equation}
    E_i = I(F(\mathcal{X}'_i)), 
\end{equation} where $I(\cdot)$ is a MOP performance indicator, and $\mathcal{X}'_i$ is the subset of solutions in $\mathcal{X}'$ that were sampled from the distribution $G(\mathbf{x}' | \bm{\mu}_i, \bm{\Sigma})$.

We consider the \textit{hypervolume indicator} \cite{brockhoff08a} (HI) as a convenient $I(\cdot)$, as it measures simultaneously dominance and diversity of solutions \cite{auger09a} and is the only indicator that is known to be strictly monotonic with regard to Pareto dominance \cite{bader11a}. The HI of $\mathcal{X}'_i$ is the hypervolume of the space that is dominated by $\mathcal{X}'_i$ and is bounded by a reference point $\bm{r} \in \mathbb{R}^m$:

\begin{equation}
    I(\mathcal{X}'_i) = \lambda^*\left(\bigcup\limits_{\mathbf{x'} \in \mathcal{X}'_i} [r_{1}, F_1(\mathbf{x}')] \times ... \times [r_m, F_m(\mathbf{x}')]\right), 
\end{equation} where $\lambda^*$ is the Lebesgue measure, and $[\cdot, \cdot] \times ... \times [\cdot, \cdot]$ is a cartesian product between closed intervals. The reference point is chosen as $\bm{r} = [\max(F_1(\mathcal{X}')), ..., \max(F_m(\mathcal{X}'))]$ such that is always dominated by all the solutions in $\mathcal{X}'$.

It is worth to mention that the high computational cost involved in computing the HI when increasing the number of objectives could limit its application. We refer the reader to \cite{falcon20a} for alternative MOP performance indicators that can be used as fitness functions.

Once the fitness values for all the individuals of the population have been calculated, a new generation of LIs is obtained by applying evolutionary operators to the population. For this purpose, the chromosome representation is obtained by direct association to the LI tuple defined in equation \ref{eq:li_structure}. 

The application of crossover and mutation operators then will depend on whether the sets of surrogates $\mathcal{S}$ and optimization algorithms $\mathcal{A}$ are defined with fixed or adjustable parameterization. For example, as a result of a crossover operation, the existing genes of two LIs (either their surrogates or optimization algorithm) will be combined; in contrast, a mutation may result in injecting a newly parameterized surrogate or optimization algorithm into the gene pool.

\subsection{PAMELI's output}
The output of the algorithm is the set of all non-dominated solutions in the dataset $\mathcal{D}$ and the best LI (the one with the highest fitness). Aligned with the NFLTs implications aforementioned, the best LI would be the result of the adaptation process to the properties of the optimization landscape of the problem, and it can be used by an analyst as a mechanism to generate new solutions estimated to be Pareto-optimal with no need of evaluating them in the actual objective space.

We remark that since in the first iteration the LIs have not been evaluated, the fittest LI is chosen as the one whose surrogate model has the lowest average approximation error, according to:

\begin{equation}
    \frac{1}{m \, |\mathcal{X}^0|} \sum_{\mathbf{x} \in \mathcal{X}^0}\sum_{j=1}^m L(\widetilde{F}_{i,j}(\mathbf{x}), F_j(\mathbf{x})). 
\end{equation}

\section{Numerical Experiments}
\label{sec:num_ex}
In this section numerical experiments are conducted to quantitatively evaluate PAMELI using the well-known DTLZ set of benchmark problems \cite{dtlz}. First, the performance indicators to support the quantitative evaluation and the experimental setup are presented. Next, we test variations of the algorithmic components of PAMELI and observe its effect in performance. Finally, we validate our proposal by comparing it against K-RVEA (see Section \ref{sec:background_saeas}), a state-of-the-art SAEA.

\subsection{Performance indicators}
We characterize the performance by means of two indicators: the \textit{Inverted Generational Distance} (IGD) \cite{igd} and a speed of convergence indicator $C$. The IGD is used to evaluate the performance at each iteration, and is computed as:

\begin{equation}
    IGD(\mathcal{G}, \mathcal{H}) =
    \frac{1}{|\mathcal{H}|}\sqrt{\sum_{j=1}^{|\mathcal{H}|}\hat{d_{j}}^2}, 
\end{equation} where $\mathcal{G} = \{\mathbf{g}_1, \mathbf{g}_2, \dots \mathbf{g}_{|G|}\}$ is the set of objective values of the obtained non-dominated solutions, $\mathcal{H} = \{\mathbf{h}_1, \mathbf{h}_2, \dots , \mathbf{h}_{|H|}\}$ represents the reference PF, and $\hat{d_{j}}$ is the Euclidean distance from $\mathbf{h}_j$ to its closest value in $\mathcal{G}$. The IGD is chosen to facilitate the comparison with other algorithms, as it is the mainly used indicator to evaluate the performance of multi-objective optimization algorithms \cite{Riquelme15a}. 

The performance across iterations is quantified in terms of the speed of convergence using the reciprocal of the area under the error curve: 

\begin{equation}
    C = \frac{1}{\int_1^{n_{iters}} e(t) \; dt}, 
    \label{eq:soc}
\end{equation} where $n_{iters}$ is the number of iterations, and the error curve $e(t)$ is obtained by linearly interpolating the IGD values between iterations. $C$ quantifies the speed of convergence since rapidly decreasing IGD values would result in a smaller area. 

\subsection{Experimental setup}

The parameters used for the PAMELI algorithm are described in Table \ref{tab:algo_params}. Besides, the pool of surrogates $\mathcal{S}$ and optimization algorithms $\mathcal{A}$ considered in this empirical study, along with their corresponding parameters are presented in Table \ref{tab:fas_params} and Table \ref{tab:heuristics_params}. Only mild assumptions are made about the objective functions, therefore $\mathcal{S}$ is composed of universal function approximators (ANNs \cite{abiodun2018}, SVRs \cite{smola2004tutorial} and FISs \cite{Mendel1995}). However, to illustrate our point of encoding proper priors, since we know beforehand that the objective functions of the DTLZ problems are non-negative, we use a ReLU function over the outputs in the ANNs and FISs. Regarding $\mathcal{A}$, its elements were chosen to include dominance-based (NSGA-II \cite{nsgaii}) and decomposition-based (MOEA/D \cite{Zhang07a}) algorithms.

For each DTLZ problem, with the parameterization showed in Table \ref{tab:probs_params}, an experiment of 30 independent runs of 10 iterations of the PAMELI algorithm is performed. As for the performance indicators, a 95\% confidence interval is calculated for the IGD using a reference set of 10000 points. We also obtain the mean and standard deviation of the speed of convergence indicator $C$. In the validation against the K-RVEA algorithm we use the parameters showed in Table \ref{tab:krvea_params} and evaluate its IGD every 100 objective function evaluations.

The implementation of PAMELI was made with the Python 3 programming language\footnote{Source code available at: \url{https://github.com/tiagoCuervo/PAMELI}}. The PlatEMO library \cite{tian2017platemo} was used for the experiments with K-RVEA and for the obtention of the reference sets employed in the calculation of the IGD.

\begin{table}[!t]
\caption{PAMELI parameters}
\centering
\begin{tabular}{@{}cc@{}}
\toprule
\toprule
Parameter & Value \\ \midrule
Population size ($n$)         & 8     \\
                              Size of the initial sample ($|\mathcal{X}^0|$)         & 100     \\
                              \makecell{Pool of surrogate models ($\mathcal{S}$)}         & \makecell{\{ANN, FIS, SVM\}}     \\
                              \makecell{Pool of optimization algorithms ($\mathcal{A}$)} &           \makecell{\{NSGA-II, MOEA/D\}}     \\
                              \makecell{Size of EPS ($|\mathcal{X}'|$)}         & 100     \\
                              Selection operator         & Roulette wheel and elitism \\
                              Crossover operator         & Uniform     \\
                              Mutation operator         & Uniform     \\
                    Probability of crossover         & 1.0     \\
                              Probability of mutation         & 0.1     \\
                              \makecell{Solutions sampled from the best LI}         & 100     \\ \bottomrule
\bottomrule
\end{tabular}
\label{tab:algo_params}
\end{table}

\begin{table}[!t]
\caption{Regression models parameters}
\centering
\begin{tabular}{@{}ccc@{}}
\toprule
\toprule
\multicolumn{1}{l}{Regression model} & Parameter & Possible values \\ \midrule
\multirow{10}{*}{ANN}          & Number of layers & 1 to 4     \\
                              & Hidden layer width & $2^a, \; a \in \{3, 4, ..., 10\}$    \\
                              & \makecell{Hidden activation function} & ReLU\\
                              & \makecell{Output activation function} & ReLU\\
                              & Optimizer & Adam \cite{Adam}\\
                              & \makecell{Learning rate} & $10^a, \; a \in \{-4, ..., -2\}$\\
                              & \makecell{Train/Validation split} & 70\%/30\% \\
                              & Stop criteria & Early stopping\\
                              & Max. epochs & 1000\\
                              \midrule
\multirow{10}{*}{FIS}          & Inference & Takagi-Sugeno    \\
                              & Number of rules & $2^a, \; a \in \{3, 4,..., 8\}$     \\
                              & \makecell{Membership function} & Gaussian    \\
                              & \makecell{Output function} & ReLU     \\
                              & Optimizer & Adam    \\
                              & \makecell{Learning rate} & $10^a, \; a \in \{-4, ..., -2\}$     \\
                              & \makecell{Train/Validation split} & 70\%/30\% \\
                              & Stop criteria & Early stopping\\
                              & Max. epochs & 1000\\
                             \midrule
\multirow{3}{*}{SVR}          & Kernel & RBF \\
& Regularization coefficient & 1.0 \\
& Kernel scale & $(d \,  \sigma^2(\mathcal{X}))^{-1}$ \\
                              \bottomrule
\bottomrule
\end{tabular}
\label{tab:fas_params}
\end{table}

\begin{table}[!t]
\caption{Optimization algorithms parameters}
\centering
\begin{tabular}{@{}ccc@{}}
\toprule
\toprule
\multicolumn{1}{l}{Algorithm} & Parameter & Possible values \\ \midrule
\multirow{4}{*}{NSGA-II}          
& Population size & 100     \\
& Number of generations & 800     \\
                              & Probability of crossover & 0.95    \\
                              & \makecell{Mutation probability} & 0.01\\
                              \midrule
\multirow{6}{*}{MOEA/D}          & Population size & 100     \\
& Number of generations & 800     \\
& Decomposition & Tchebycheff \\
& Weight’s neighborhood & 20 \\
                              & Crossover parameter & 0.95    \\
                              & Differential evolution parameter & 0.5 \\
                              \bottomrule
\bottomrule
\end{tabular}
\label{tab:heuristics_params}
\end{table}

\begin{table}[!t]
\caption{Problems parameters}
\centering
\begin{tabular}{@{}ccc@{}}
\toprule
\toprule
\thead{Problem} & \thead{Number of \\decision variables ($d$)} & \thead{Number of \\objectives ($m$)} \\ 
\midrule
DTLZ1 & 7 & 3 \\ 
DTLZ2-DTLZ6 & 12 & 3 \\ 
DTLZ7 & 22 & 3 \\ 
\bottomrule
\bottomrule
\end{tabular}
\label{tab:probs_params}
\end{table}

\begin{table}[!t]
\caption{K-RVEA parameters}
\centering
\begin{tabular}{@{}cc@{}}
\toprule
\toprule
Parameter & Value \\ \midrule
Population size         & 100     \\
                              Rate of change penalty ($\alpha$)        & 2     \\
                              \makecell{Number of iterations before updating Krigging models ($w_{max}$)}         & 20     \\
                              \makecell{Number of re-evaluated solutions at each generation ($\mu$)}         & 5     \\ \bottomrule
\bottomrule
\end{tabular}
\label{tab:krvea_params}
\end{table}

\subsection{Experiments with PAMELI's algorithmic components}
\label{sec:res_effect}

\begin{figure*}[!t]
\centering
\includegraphics[width=0.32\textwidth, height=2.5cm]{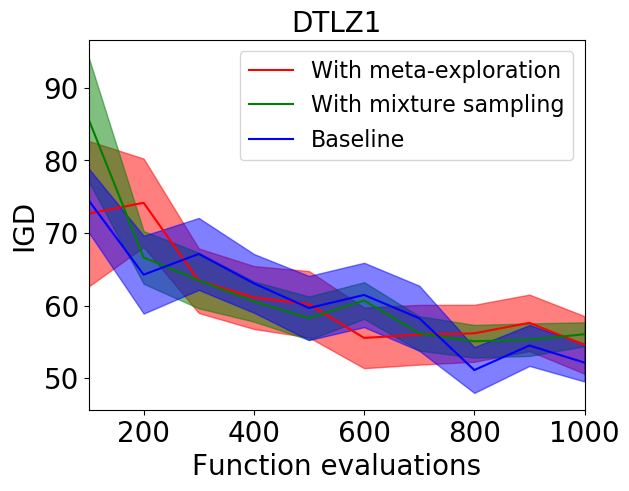}
\includegraphics[width=0.32\textwidth, height=2.5cm]{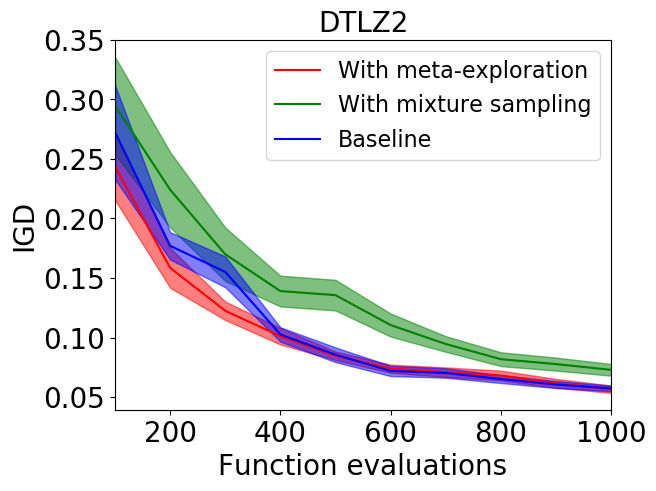}
\includegraphics[width=0.32\textwidth, height=2.5cm]{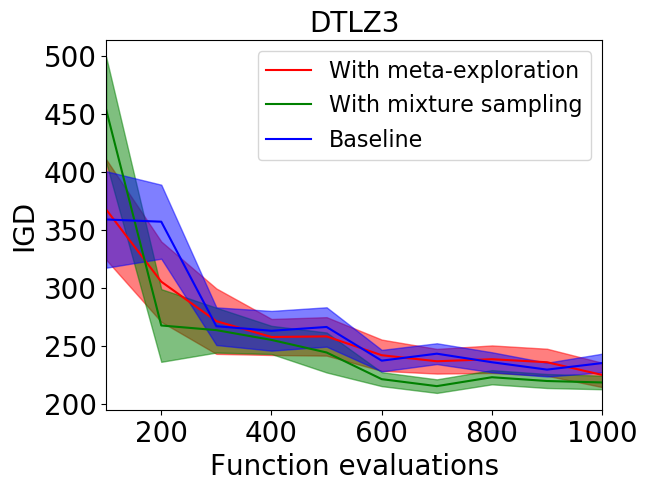}
\includegraphics[width=0.32\textwidth, height=2.5cm]{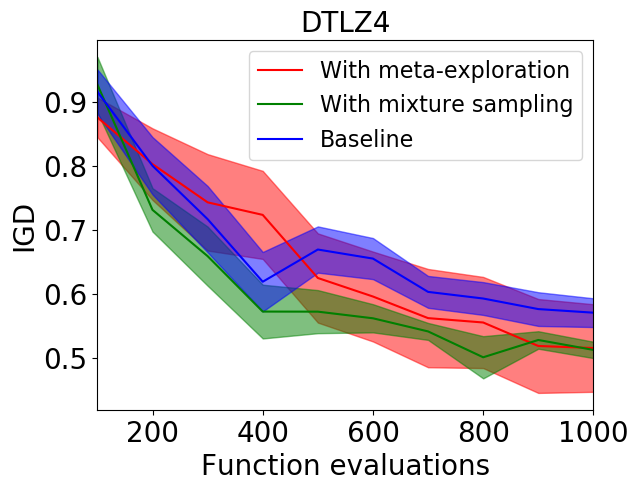}
\includegraphics[width=0.32\textwidth, height=2.5cm]{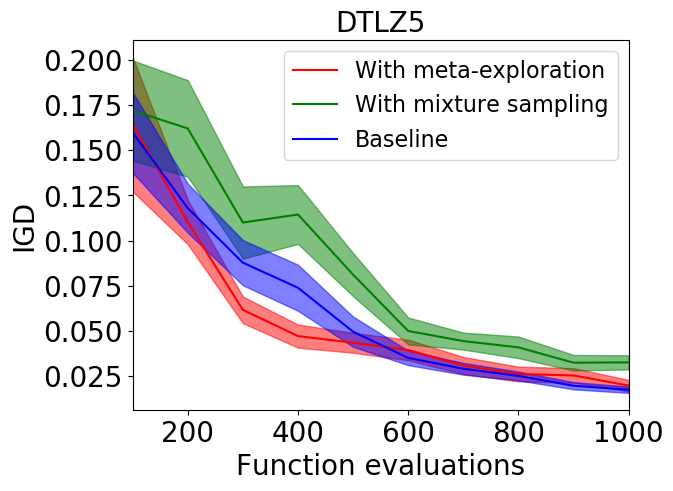}
\includegraphics[width=0.32\textwidth, height=2.5cm]{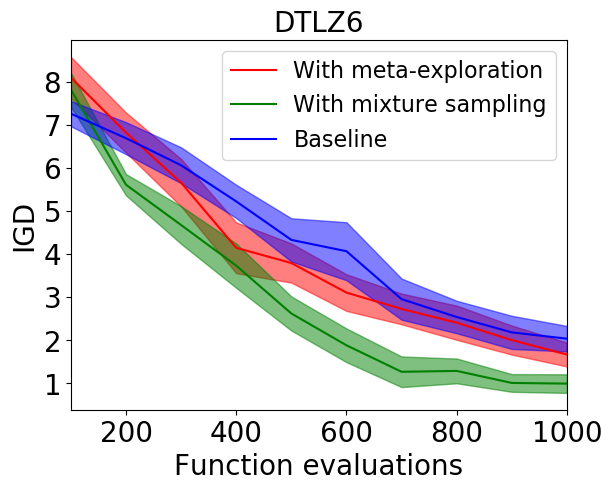}
\includegraphics[width=0.32\textwidth, height=2.5cm]{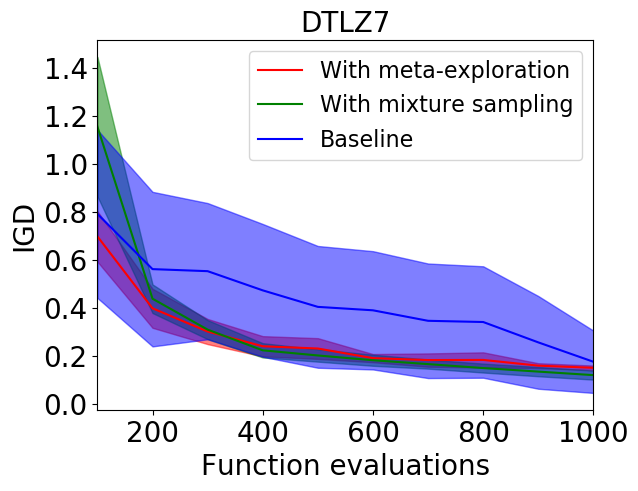}
\caption{IGD over iterations for the baseline system and its variations with adaptation of LIs or adaptative exploration. The solid line is the sample mean obtained for 30 independent runs. The shaded area shows a confidence interval of 99.5\%.}
\label{fig:pamelicomp_dtlz}
\end{figure*}

To assess the effect of the proposed exploration strategy to build the EPS and the meta-exploration for optimal LIs, we compared three different versions of the algorithm. A \emph{baseline} version, differing from the algorithm described in Section \ref{sec:algorithm} in that it uses a constant population of LIs after initialization and samples the promising solutions directly from the APSs with uniform probability; as a result, it favors a highly exploitative strategy. The second version resorts to \emph{only meta-exploration} (OME) as it also employs the exploitative sampling strategy to evolve the population of LIs. And finally, we tried a third version adopting \emph{only mixture sampling} (OMS), that uses the sampling strategy based in the mixture distribution described in Section \ref{sec:algorithm}, but does not perform meta-exploration.

\begin{table*}[]
\caption{Mean IGD of the final solutions}
\centering
\setlength{\tabcolsep}{2.2pt}
\begin{tabular}{ccccccccc}
\toprule
& \multicolumn{1}{c}{DTLZ1}   & \multicolumn{1}{c}{DTLZ2}  & \multicolumn{1}{c}{DTLZ3}     & \multicolumn{1}{c}{DTLZ4}  & \multicolumn{1}{c}{DTLZ5}  & \multicolumn{1}{c}{DTLZ6}  & \multicolumn{1}{c}{DTLZ7}  \\ \midrule
Baseline                                                                          & $\mathbf{51.100 \pm 8.449}$ & $0.057 \pm 0.009$          & $229.000 \pm 15.894$          & $0.571 \pm 0.070$          & $\mathbf{0.018 \pm 0.005}$ & $2.037 \pm 0.779$          & $0.176 \pm 0.342$          \\ \cline{1-8} 
                                OME  & $54.600  \pm 10.631$        & $\mathbf{0.057 \pm 0.008}$ & $225.000 \pm 28.061$          & $0.516 \pm 0.183$          & $0.020 \pm 0.008$          & $1.670 \pm 0.730$          & $0.151 \pm 0.025$          \\ \cline{1-8} 
                                OMS & $55.100 \pm 6.348$          & $0.073 \pm 0.013$          & $\mathbf{215.000 \pm 15.467}$ & $\mathbf{0.502 \pm 0.036}$ & $0.032 \pm 0.016$          & $\mathbf{0.996 \pm 0.569}$ & $\mathbf{0.120 \pm 0.049}$ \\ \cline{1-8} 
\end{tabular}
\label{tab:Mean_IGD_One}
\end{table*}

The three versions were tested on the DTLZ problem set. The obtained IGD curves are depicted in Fig.\ref{fig:pamelicomp_dtlz}, whereas in Table \ref{tab:Mean_IGD_One} the best IGD mean is presented. The speed of convergence indicator $C$ for each algorithm is shown in Table \ref{tab:pamelicomp_dtlz}. Below we highlight some important observations:

\begin{itemize}
    \item In almost all problems the baseline is outperformed in terms of speed of convergence by at least one of the other versions except for DTLZ1 and DTLZ7. It obtained the best mean value in DTLZ4, but with large deviations relative to the other versions. 
    
    \item Regarding the quality of the final solutions, the baseline presented the best performance in DTLZ1, and DTLZ5.
    
    \item OMS obtained the lowest mean IGD in DTLZ3, DTLZ4, DTLZ6 and DTLZ7, outperforming the baseline.

    \item Concerning to the speed of convergence, OMS presented the highest values with respect to the baseline in DTLZ4 and DTLZ6. 

    \item OME resulted in faster convergence in DTLZ2 and DTLZ5. 
    
    \item In terms of the quality of the final solutions OME is outperformed in almost all problems by at least one of the other versions, except for DTLZ2.
\end{itemize}

\begin{table*}[]
\caption{Speed of convergence indicator ($C$) for the baseline algorithm and its variations with adaptative exploration and adaptative landscape identifiers on the DTLZ problems}
\centering
\setlength{\tabcolsep}{2 pt}
\begin{tabular}{cccccccc}
\toprule
                                             & \multicolumn{1}{c}{DTLZ1} & \multicolumn{1}{c}{DTLZ2} & \multicolumn{1}{c}{DTLZ3} & \multicolumn{1}{c}{DTLZ4} & \multicolumn{1}{c}{DTLZ5} & \multicolumn{1}{c}{DTLZ6} & \multicolumn{1}{c}{DTLZ7} \\  \midrule 
 Baseline                                          & $\mathbf{0.0019 \pm0.0002}$       & $1.0722\pm0.1546$       & $\mathbf{0.0004\pm0.0000}$       & $0.1696\pm0.0193$       & $2.0475\pm0.5460$       & $0.0269\pm0.0057$       & $\mathbf{1.0160\pm0.7285}$       \\ \cline{1-8} 
                      {\makecell{OME}} & $0.0019\pm0.0003$       & $\mathbf{1.1417\pm0.1541}$       & $0.0004\pm0.0001$       & $0.1830\pm0.0475$       & $\mathbf{2.2223\pm0.4575}$       & $0.0299\pm0.0079$       & $0.4704\pm0.1190$       \\  \cline{1-8} 
                     {\makecell{OMS}} & $0.0018\pm0.0002$       & $0.8590\pm0.1767$       & $\mathbf{0.0004\pm0.0000}$       & $\mathbf{0.1872\pm0.0177}$       & $1.4016\pm0.2497$       & $\mathbf{0.0405\pm0.0109}$       & $0.4432\pm0.1444$       \\ \cline{1-8} 
\end{tabular}
\label{tab:pamelicomp_dtlz}
\end{table*}

\begin{figure*}[!ht]
\centering
\includegraphics[width=0.32\textwidth, height=2.5cm]{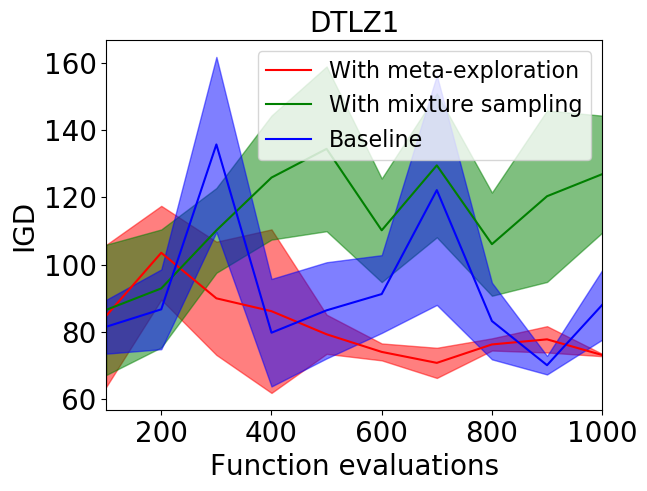}
\includegraphics[width=0.32\textwidth, height=2.5cm]{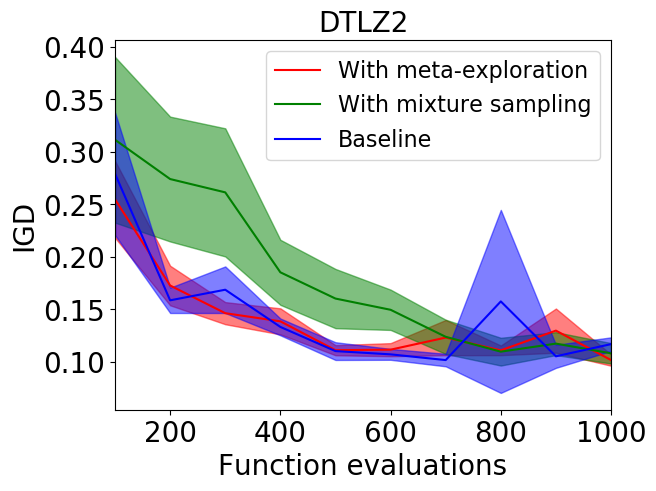}
\includegraphics[width=0.32\textwidth, height=2.5cm]{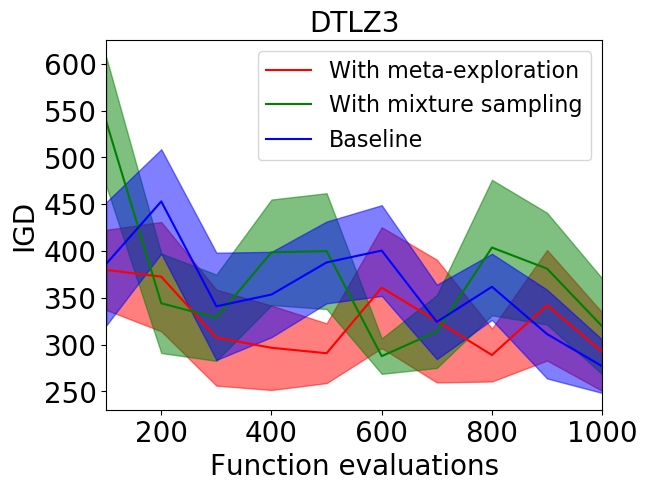}
\includegraphics[width=0.32\textwidth, height=2.5cm]{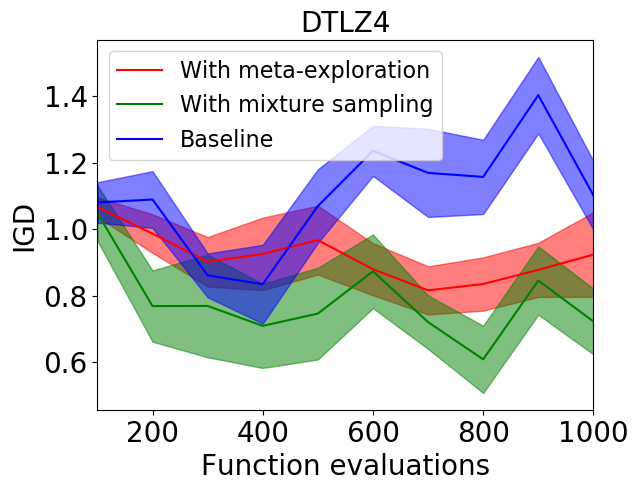}
\includegraphics[width=0.32\textwidth, height=2.5cm]{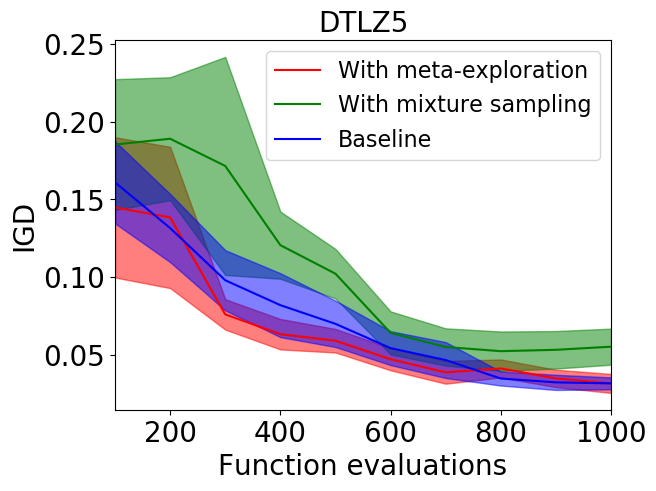}
\includegraphics[width=0.32\textwidth, height=2.5cm]{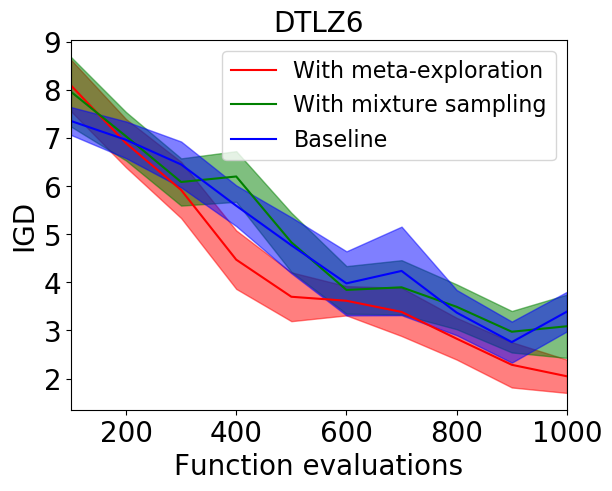}
\includegraphics[width=0.32\textwidth, height=2.5cm]{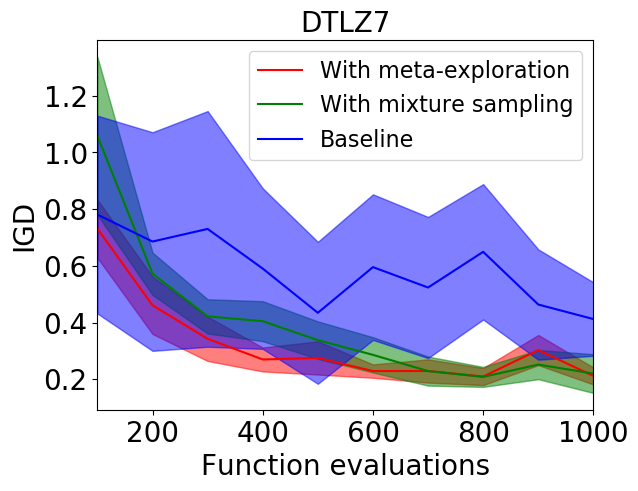}
\caption{IGD over iterations calculated with the $\mathcal{X}_f$ set for the baseline system and its variations with adaptation of LIs or adaptative exploration. The solid line is the sample mean obtained for 30 independent runs. The shaded area shows a confidence interval of 99.5\%.}
\label{fig:approxcomp_dtlz}
\end{figure*}

We recall that being one of the output of PAMELI, the evolved best LI can be used to sample new estimated Pareto-optimal solutions. Therefore we also evaluated the quality of only those samples generated with the resulting LIs across all iterations. Fig.\ref{fig:approxcomp_dtlz} depicts the obtained IGD curves and Table \ref{tab:approxcomp_dtlz} reports the values of the speed of convergence indicator $C$, showing that:

\begin{itemize}
    \item OME shows better performance in all problems except DTLZ4 and DTLZ7, where it is outperformed by OMS and baseline respectively.
    \item In DTLZ1 OME is the only version that exhibits convergence but does not show a clear improving trend across iterations.
    \item In DTLZ4 the solutions sampled with the LIs do not exhibit a clear IGD improving trend across iterations. 
\end{itemize}

\begin{table*}[]
\caption{Speed of convergence indicator ($C$) evaluated on the $\mathcal{X}_f$ set for the baseline algorithm and its variations with adaptative exploration and adaptative landscape identifiers on the DTLZ problems}
\centering
\setlength{\tabcolsep}{2.2pt}
\begin{tabular}{ccccccccc}
\toprule
                & \multicolumn{1}{c}{DTLZ1}              & \multicolumn{1}{c}{DTLZ2}              & \multicolumn{1}{c}{DTLZ3}              & \multicolumn{1}{c}{DTLZ4}              & \multicolumn{1}{c}{DTLZ5}              & \multicolumn{1}{c}{DTLZ6}              & \multicolumn{1}{c}{DTLZ7}              \\ \midrule
Baseline              & $0.0013\pm0.0002$ & $0.8279\pm0.1578$ & $0.0003\pm0.0001$ & $0.1020\pm0.0107$ & $1.7133\pm0.4945$ & $0.0236\pm0.0040$ & {$\mathbf{0.7469\pm0.6985}$ }
    \\ \cline{1-9} 
    {
    }OME &      {$\mathbf{0.0014\pm0.0002}$} & {$\mathbf{0.8359\pm0.0963}$} & {$\mathbf{0.0004\pm0.0001}$} & $0.1263\pm0.0233$ & {$\mathbf{1.9231\pm0.5445}$} & {$\mathbf{0.0274\pm0.0060}$} & $0.3855\pm0.0855$ 
    \\ \cline{1-9} {
    }OMS & $0.0010\pm0.0001$ & $0.6928\pm0.1845$ & $0.0003\pm0.0001$ & {$\mathbf{0.1504\pm0.0301}$} & $1.1460\pm0.2548$ & $0.0234\pm0.0040$ & $0.3404\pm0.1343$ \\ \cline{9-9}
    \bottomrule
\end{tabular}
\label{tab:approxcomp_dtlz}
\end{table*}

\subsection{Validation}

\begin{table*}[]
\caption{Comparison of the speed of convergence indicator ($C$) between K-RVEA and PAMELI on the DTLZ problems}
\setlength{\tabcolsep}{2pt}
\centering
\begin{tabular}{ccccccccc}
\toprule
\multicolumn{1}{c}{}           & \multicolumn{1}{c}{DTLZ1} & \multicolumn{1}{c}{DTLZ2} & \multicolumn{1}{c}{DTLZ3} & \multicolumn{1}{c}{DTLZ4} & \multicolumn{1}{c}{DTLZ5} & \multicolumn{1}{c}{DTLZ6} & \multicolumn{1}{c}{DTLZ7} \\ \midrule
K-RVEA 
& $\mathbf{0.0021\pm0.0004}$       
& $0.7345\pm0.0806$       & $0.0003\pm0.0000$       & $\mathbf{0.3322\pm0.0463}$       
& $0.7135\pm0.0684$       & $0.0236\pm0.0011$       & $0.1741\pm0.0304$       \\ \cline{1-8}
PAMELI 
& $0.0019\pm0.0004$       & $\mathbf{0.9470\pm0.1947}$       
& $\mathbf{0.0004\pm0.0000}$       
& $0.1908\pm0.0407$       & $\mathbf{1.6788\pm0.3760}$       
& $\mathbf{0.0396\pm0.0072}$       & $\mathbf{0.6826\pm0.2519}$       \\ \bottomrule 
\end{tabular}
\label{tab:compalgos_dtlz}
\end{table*}

\begin{figure*}[!ht]
\centering
\includegraphics[width=0.32\textwidth, height=2.5cm]{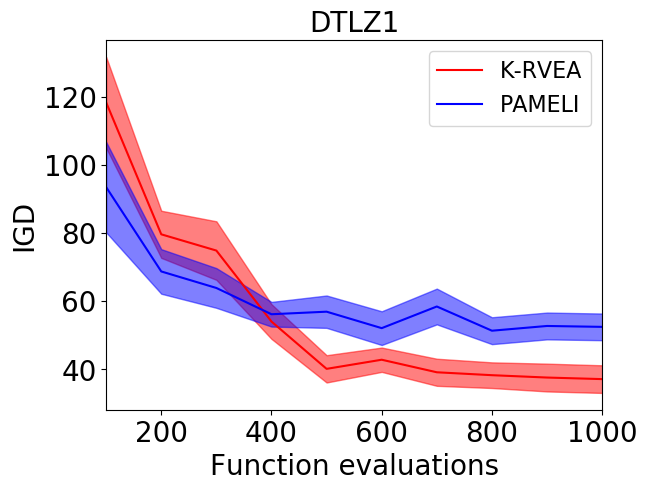}
\includegraphics[width=0.32\textwidth, height=2.5cm]{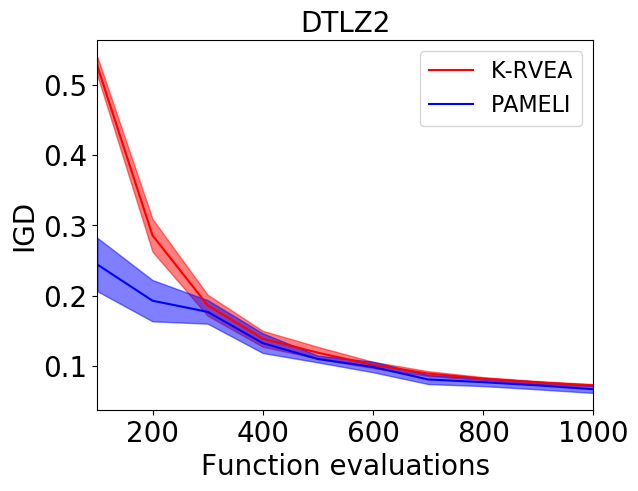}
\includegraphics[width=0.32\textwidth, height=2.5cm]{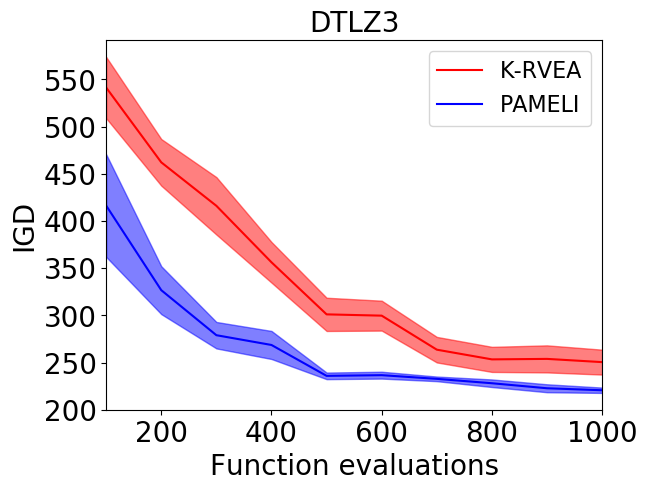}
\includegraphics[width=0.32\textwidth, height=2.5cm]{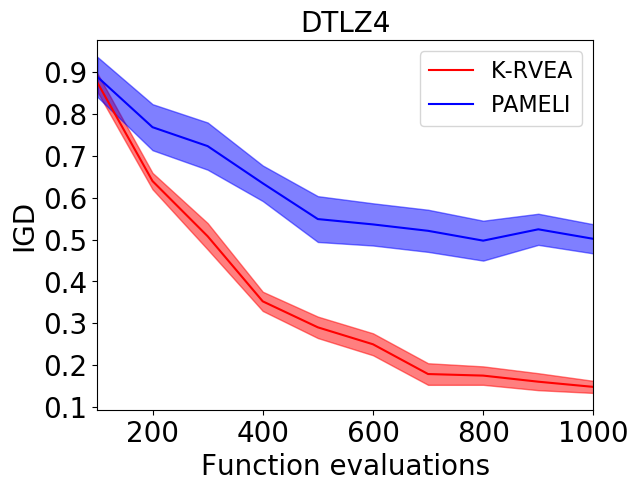}
\includegraphics[width=0.32\textwidth, height=2.5cm]{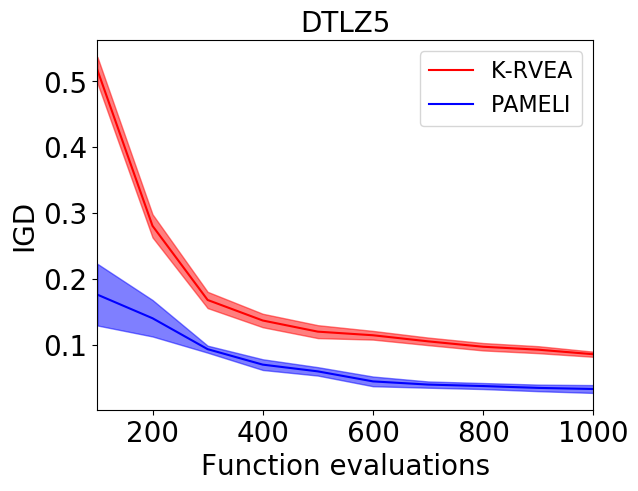}
\includegraphics[width=0.32\textwidth, height=2.5cm]{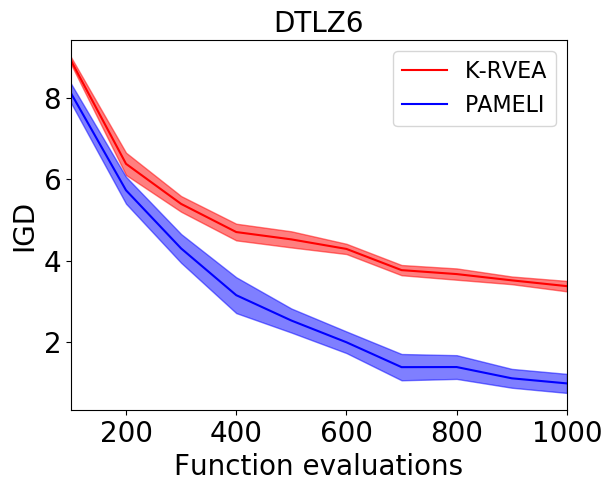}
\includegraphics[width=0.32\textwidth, height=2.5cm]{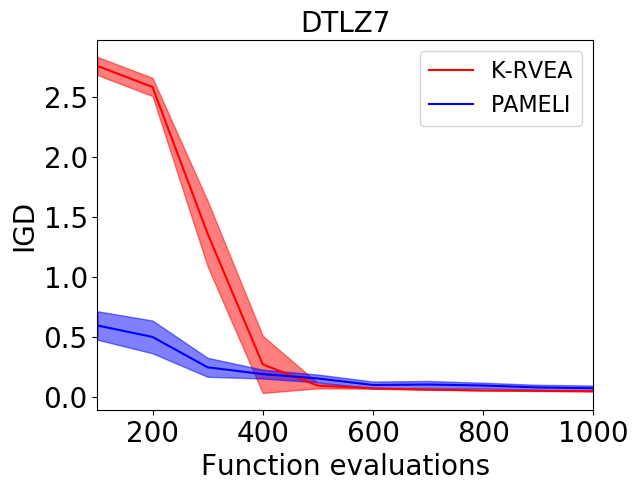}
\caption{IGD over iterations for PAMELI and K-RVEA on the DTLZ problem set. The solid line is the sample mean obtained for 30 independent runs. The shaded area shows a confidence interval of 99.5\%.}
\label{fig:compalgos_dtlz}
\end{figure*}

\begin{figure*}[!ht]
\centering
\includegraphics[width=0.32\textwidth, height=2.7cm]{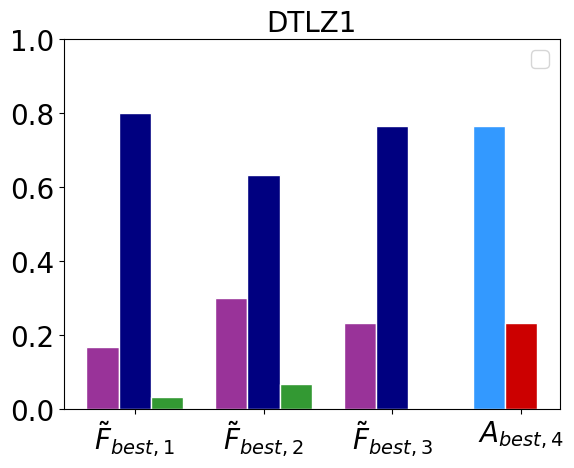}
\includegraphics[width=0.32\textwidth, height=2.7cm]{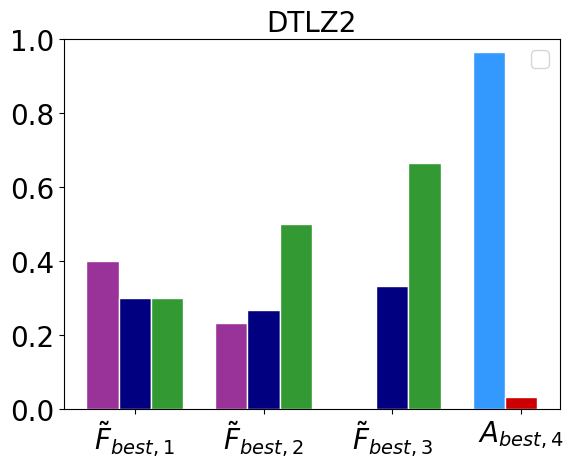}
\includegraphics[width=0.32\textwidth, height=2.7cm]{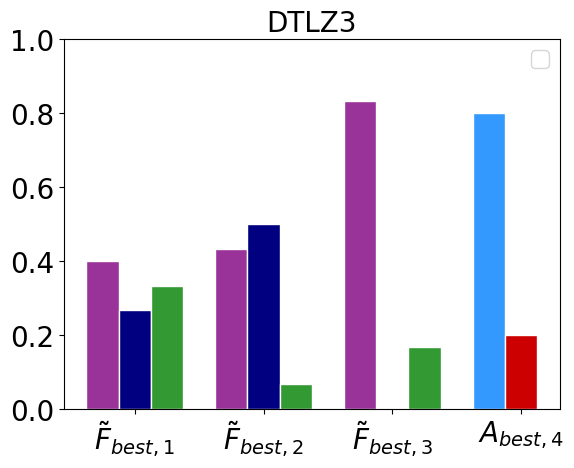}
\includegraphics[width=0.32\textwidth, height=2.7cm]{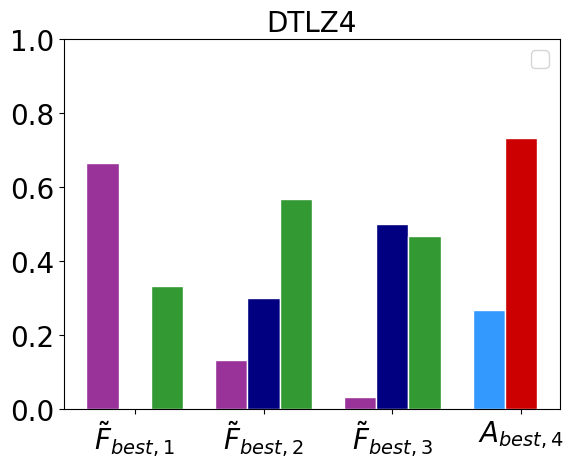}
\includegraphics[width=0.32\textwidth, height=2.7cm]{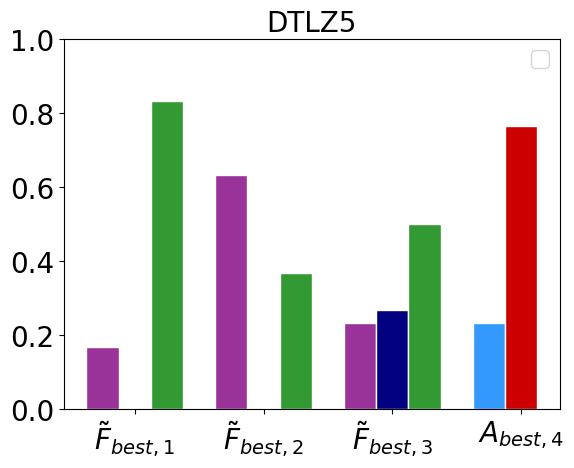}
\includegraphics[width=0.32\textwidth, height=2.7cm]{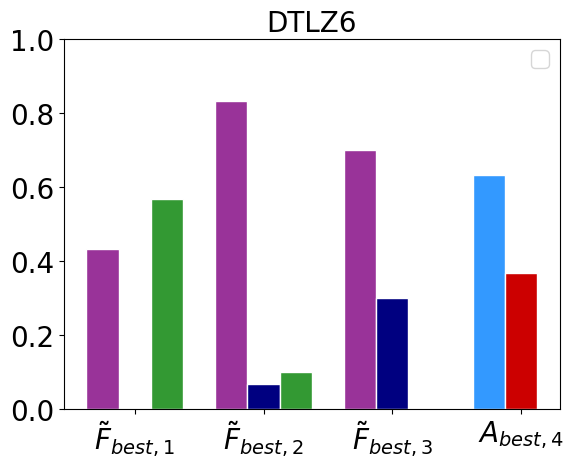}
\includegraphics[width=0.425\textwidth, height=2.7cm]{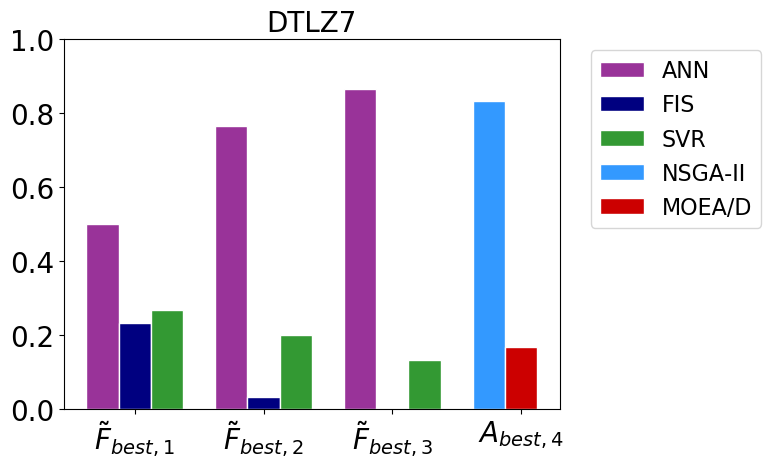}
\caption{Bar charts illustrating the obtained distribution of the structure of the best LIs for the 30 independent runs. Regarding the surrogate models, only the class of regression method was considered for the count, regardless of its parameters.}
\label{fig:barchart_LIs}
\end{figure*}
\label{sec:pameli_vs_krvea}

\begin{table*}[]
\caption{Mean IGD of the final solutions with the best LI}
\centering
\setlength{\tabcolsep}{4.5pt}
\begin{tabular}{cccccccc}
\toprule
\multicolumn{1}{c}{}  & \multicolumn{1}{c}{DTLZ1} & \multicolumn{1}{c}{DTLZ2} & \multicolumn{1}{c}{DTLZ3} & \multicolumn{1}{c}{DTLZ4} & \multicolumn{1}{c}{DTLZ5} & \multicolumn{1}{c}{DTLZ6} & \multicolumn{1}{c}{DTLZ7} \\ \midrule
 K-RVEA               
 & $\mathbf{37.200 \pm 26.811}$
 & $0.073 \pm 0.142$      & $250.000 \pm 101.805$  & $\mathbf{0.148 \pm 0.244}$      
 & $0.086 \pm 0.134$      & $3.380 \pm 1.701$      & $\mathbf{0.050 \pm 1.097}$         \\ \cline{1-8}
PAMELI
& $51.300 \pm 12.811$     & $\mathbf{0.067 \pm 0.060}$       
& $\mathbf{221.000 \pm 61.949}$    
& $0.497 \pm 0.135$       
& $\mathbf{0.033 \pm 0.049}$       
& $\mathbf{0.990 \pm 2.338}$       
& $0.075 \pm 0.186$         \\ \bottomrule
\end{tabular}
\label{tab:Mean_IGD_One_BestLI}
\end{table*}

Results of PAMELI and K-RVEA in the DTLZ problem set are depicted in Fig.\ref{fig:compalgos_dtlz} and Tables \ref{tab:compalgos_dtlz} and \ref{tab:Mean_IGD_One_BestLI}. It can be seen that:

\begin{itemize}
    \item In DTLZ2 $(28.93)\%$, DTLZ3 $(33.33)\%$, DTLZ5 $(135.29)\%$, DTLZ6 $(67.79)\%$ and DTLZ7 $(292.07)\%$ PAMELI outperformed K-RVEA in terms of speed of convergence.
    \item PAMELI achieved better performance in the quality of solutions in DTLZ2 $(7.69)\%$, DTLZ3 $(11.6)\%$, DTLZ5 $(61.68)\%$ and DTLZ6 $(70.71)\%$ problems.
    \item PAMELI showed a lower initial IGD in almost all problems except in DTLZ4, where K-RVEA exhibited a lower value $(1.6)\%$. The percentage differences in each problem were: DTLZ1 $(20.84)\%$, DTLZ2 $(53.61)\%$, DTLZ3 $(23.11)\%$, DTLZ5 $(65.96)\%$, DTLZ6 $(8.98)\%$ and DTLZ7 $(78.33)\%$,
    \item Regarding the quality of final solutions K-RVEA outperformed PAMELI in DTLZ1 $(37.9)\%$, DTLZ4 $(235.81)\%$ and DTLZ7 $(49.6)\%$.
    \item In DTLZ1, DTLZ3 and DTLZ6 the resulting high IGD values showed that neither PAMELI nor K-RVEA converged to the true PF.
\end{itemize}

To observe the effect of adaptation of the LIs we present in Figure \ref{fig:barchart_LIs} the relative appearance frequency of the considered regression methods and optimization algorithms in the best resulting LIs. Only the used algorithm was considered for the counting, regardless of its parameters. From these results it can be noticed that:

\begin{itemize}
    \item No single regression method is favored in all problems but at least one regression method is preferred in each function.
    \item NSGA-II appears more frequently in all problems with the exception of DTLZ4 and DTLZ5, where MOEA/D is the favored optimization algorithm.
\end{itemize}

\section{Discussion}
\label{sec:discussion}
In this section we first discuss the effects of the algorithmic components and the validation experiments between PAMELI and K-RVEA (Section \ref{sec:res_analysis}). Then, we examine the advantages exhibited by PAMELI on the empirical results (\ref{sec:observed_features}) and lastly, we reflect on some of the observed weaknesses and suggest possible ways to address them (Section \ref{sec:to_try}). 

\subsection{Algorithmic and validation experiments}\label{sec:res_analysis}

The experiments with PAMELI's algorithmic components allowed us to assess the effect of the meta-exploration and mixture model based sampling of promising solutions in the performance of the algorithm. The observations in Section \ref{sec:res_effect} suggest that meta-exploration has significant benefits regarding the number of objective function evaluations required for convergence, as it accelerated convergence in most problems, and likewise, it accelerated the improvement of the LIs.

Moreover, the information sharing mechanism associated to the mixture model version of the algorithm excelled on DTLZ3, DTLZ6, DTLZ4 and DTLZ7, which are arguably problems demanding a wider, scattered exploration, since they either have many local optima (DTLZ3 and DTLZ6), non-uniform density of Pareto optimal solutions (DTLZ4), or disconnected sets of Pareto-optimal regions (DTLZ7).

On the other hand, in DTLZ1, DTLZ3 and DTLZ4 none of the algorithms converged to the true PS, as evidenced by the high IGD values in Fig.\ref{fig:pamelicomp_dtlz}. This implies that none of the considered LIs is a good fit for these problems.

Regarding the results in terms of the final mean IGD, it can be noticed that neither meta-exploration or mixture model based sampling seem to have a significant effect. With enough iterations all versions tend to converge to similar IGD values but they do so at different speeds.

The results presented in Tables \ref{tab:Mean_IGD_One_BestLI} and \ref{tab:pamelicomp_dtlz} show that the union of meta-exploration and the mixture based sampling strategy in PAMELI enables the algorithm to benefit from both of these components as it presents a similar performance to the best of the versions in each problem.

Now, PAMELI outperformed K-RVEA on most of DTLZ problems in the validation experiments, in terms of speed of convergence, hinting at the convenience of PAMELI's mechanisms as an alternative to the  dominant SAEAs paradigm.

Besides, despite that in DTLZ2, DTLZ5 and DTLZ7 the initial IGD was lower for PAMELI, it is worth noting that starting from the dataset of the very first iteration, the best LI provides a set of solutions estimated to be Pareto-optimal, which for these problems serves as an educated guess enough to achieve a well-performing initial estimate of the PS.

K-RVEA outperformed PAMELI in DTLZ1 and DTLZ4 that, as mentioned in section \ref{sec:res_effect}, were {experiments whose evolved LIs} seemed not to be well suited. In DTLZ1, DTLZ3 and DTLZ6 the high IGD values show that neither converged to the true PF, indicating that both algorithms may become trapped in local minima. 

Lastly, let us remark on the evolved structure of the best LIs. Since the distribution of appearance of its constituent regression and optimization components resulted in a heterogeneous mix, one can infer that PAMELI is adapting the structure of the LIs to the particular properties of the problem landscape, with no bias towards any of the models in the pool of surrogates or in optimization algorithm set.

\subsection{Observed features of PAMELI}
\label{sec:observed_features}

The experimental results seem to suggest that it is the interaction between the components of PAMELI rather than its individual contributions what it gives rise to its overall superior performance. Therefore, we now highlight the features that we consider are the most relevant of our proposal and relate them to the observed performance:

\begin{itemize}
    \item In line with the conclusions of the NFLTs, PAMELI strives for specialization rather than generalization, since the meta-exploration for optimal LIs adapts the underlying optimization algorithm to the properties of the problem using the information it acquires as a byproduct of the search of solutions. 

    \item The output of PAMELI is richer than that of other algorithms, as it not only provides a set of solutions, but also a mechanism to generate any desired amount of solutions that are estimated to be optimal, therefore allowing to keep exploring the promising regions of the optimization landscape without incurring in additional objective function evaluations. 
    
    \item By using the covariance matrix computed over the samples in the union of all APSs the algorithm balances exploration and exploitation as it adapts the amount of exploration to the uncertainty in the APSs. If there is consensus (i.e. the LIs predict the same value for the PS) between the LIs then the sample variance is minimal, and hence, the region of the search space to be evaluated in the real objective space is narrow. On the contrary, if there is high variance in the predictions, meaning high uncertainty, then the algorithm explores a wider region of the search space.
\end{itemize}

\subsection{To try further}
\label{sec:to_try}
Despite the mentioned advantages, there are several likely ways in which the meta-algorithm could be improved. For instance, on DTLZ4 we identified a deficiency with problems with a non-uniform density of Pareto optimal solutions. This issue could be countered by considering reference-vector based algorithms as part of the candidate optimization algorithms to promote uniformly distributed solutions on the PF. In general, considering a larger pool of LIs could also help to alleviate the weak performance observed in DTLZ1, DTLZ3 and DTLZ4, as with more diverse surrogate models and optimization algorithms is more probable to find suitable LIs to a broader class of problems. PAMELI can use existing or new optimization algorithms, which would be a kind of assimilation where a heuristic takes advantage of the features of other ones. Thus PAMELI can be developed more like an open framework for evolutionary computing rather than a simple algorithm.

Besides, it is worth to mention that in the performed experiments not many a-priori assumptions were made about the problems. Assumptions about the optimization landscapes (e.g. linearity, convexity, continuity) and about the properties related to the multi-objective nature of the problem (e.g. decomposability, deceptiveness of the PF, isolated optima), can, and should be taken into account where possible by appropriately specifying the set of candidate LIs. Considering that the more knowledge that can be inserted a priori into the algorithm, the better, this could also be a possible way of improvement.
Additionally, solving problems with high-dimensional objective spaces could imply challenges, as the hypervolume-based fitness function becomes computationally expensive. The genetic algorithm could also be unfit for high-dimensional problems, given that as the number of objectives increases,  the search space of the meta-search for LIs increases too, and it may require more iterations and larger populations to achieve good solutions. A possible way to address this issues is to replace the genetic algorithm with a different selection strategy to guide the search process for LIs. For instance, the LIs could be selected based on a priori knowledge about its suitability to the optimization landscape at hand. This process could be carried out based on expert knowledge or on computational techniques \cite{Mersmann11a}. 

Likewise, there are some changes to the actual instance of PAMELI that are worth exploring. The concept of LI could be extended by considering the construction of the EPS as part of the landscape identification process. This would broaden the search space of the meta-exploration, as it would include probabilistic models aside of surrogate models and optimization algorithms. Moreover, the local search that the meta-exploration performs could exploit promising regions of the space of LIs by using a mutation operator that modifies the parameters of the components of the LIs.

Finally, it should be noted that the meta-algorithm may incur in a high computational burden by training multiple surrogate-models and solving multiple optimization problems in each iteration; however, compared to the cost of using the real objective function, it may seem negligible. Besides, the algorithm is highly parallelizable, since the training of the surrogate model and the estimation of the PS (inner loop in algorithm \ref{algo:pameli}), that is the most expensive step in the algorithm, can be done concurrently for each LI.

\section{Conclusions}
\label{sec:conclusions}
In this paper we have presented PAMELI, an algorithm for finding approximate solutions to computationally expensive multi-objective optimization problems. Our main contribution is a meta-algorithm that adapts the components of an underlying optimization process to better suit the properties of the optimization landscape at hand, which showed to improve its performance in terms of the number of objective function evaluations required for convergence to the PF. This strategy goes in line with the consequences of the NFLTs, by striving for an algorithm that specializes to the problem as it explores it, rather than attempting to generalize. 
 
In essence, PAMELI performs a meta-exploration of Landscape Identifiers, which we define as a combination of a surrogate model and an optimization algorithm (so as to approximate one function objective and to sample the actual function, respectively). One key feature of the algorithm is that the real problem landscape guides the evolution of the most suitable LIs. Furthermore, the LIs can be configured with prior information on the properties of the objective functions, providing PAMELI with wide versatility to adapt to different types of problem landscapes. 
 
The presented algorithm has several desirable features. For each  set of evaluations on the real objective space, PAMELI produces a set of solutions estimated to be Pareto-optimal, therefore making the most of the available data. Moreover, surrogate models of the objective functions are used to save expensive computation of additional evaluations. The surrogate models are also used to balance the exploration/exploitation trade-off, by adjusting the scope of the search for solutions to the uncertainty in the models about the PS. Furthermore, PAMELI can acquire the features of any other multi-objective optimization algorithm by considering it as a candidate optimization algorithm in the pool of LIs.

The empirical results seem to indicate the suitability of PAMELI as a meta-algorithm for adaptive navigation of optimization landscapes. The experiments performed on the DTLZ problems comparing PAMELI against K-RVEA showed that in most problems PAMELI achieves better solutions with less function evaluations. In the future we would like to compare the performance of PAMELI with HSMEA, as it reports state-of-the art performance. This was not possible in this work as no available implementation of HSMEA was found. 

Finally, the question of how well PAMELI can work on other families of optimization problems, such as single-objective and many-objective optimization (problems with more than three goals) is considered an interesting avenue for future work. 

\section*{Acknowledgment}

We would like to thank the \textit{Laboratory of Automation and Computational Intelligence} (LAMIC) from the \textit{Universidad Distrital Francisco José de Caldas} for their support during the development of this project.

\ifCLASSOPTIONcaptionsoff
  \newpage
\fi



\bibliographystyle{IEEEtran}
\bibliography{main}
%



%

\begin{IEEEbiography}[{\includegraphics[width=1in,height=1.25in,clip,keepaspectratio]{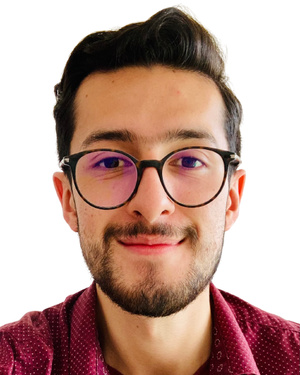}}]{Santiago Cuervo}
received the BSc degree in electronics engineering from the Universidad Distrital Francisco José de Caldas, Bogot\'a, Colombia, in 2019. He was awarded the Ignacy Łukasiewicz Scholarship by the government of the Republic of Poland, and since 2020 is a student of the Master in Data Science in the University of Wroclaw, Wroclaw, Poland. His current research interests are on deep learning for agent-based problems, and the application of non-euclidean deep learning to the analysis of optimization landscapes.
\end{IEEEbiography}

\vspace{-.5cm}
\begin{IEEEbiography}[{\includegraphics[width=1in,height=1.25in,clip,keepaspectratio]{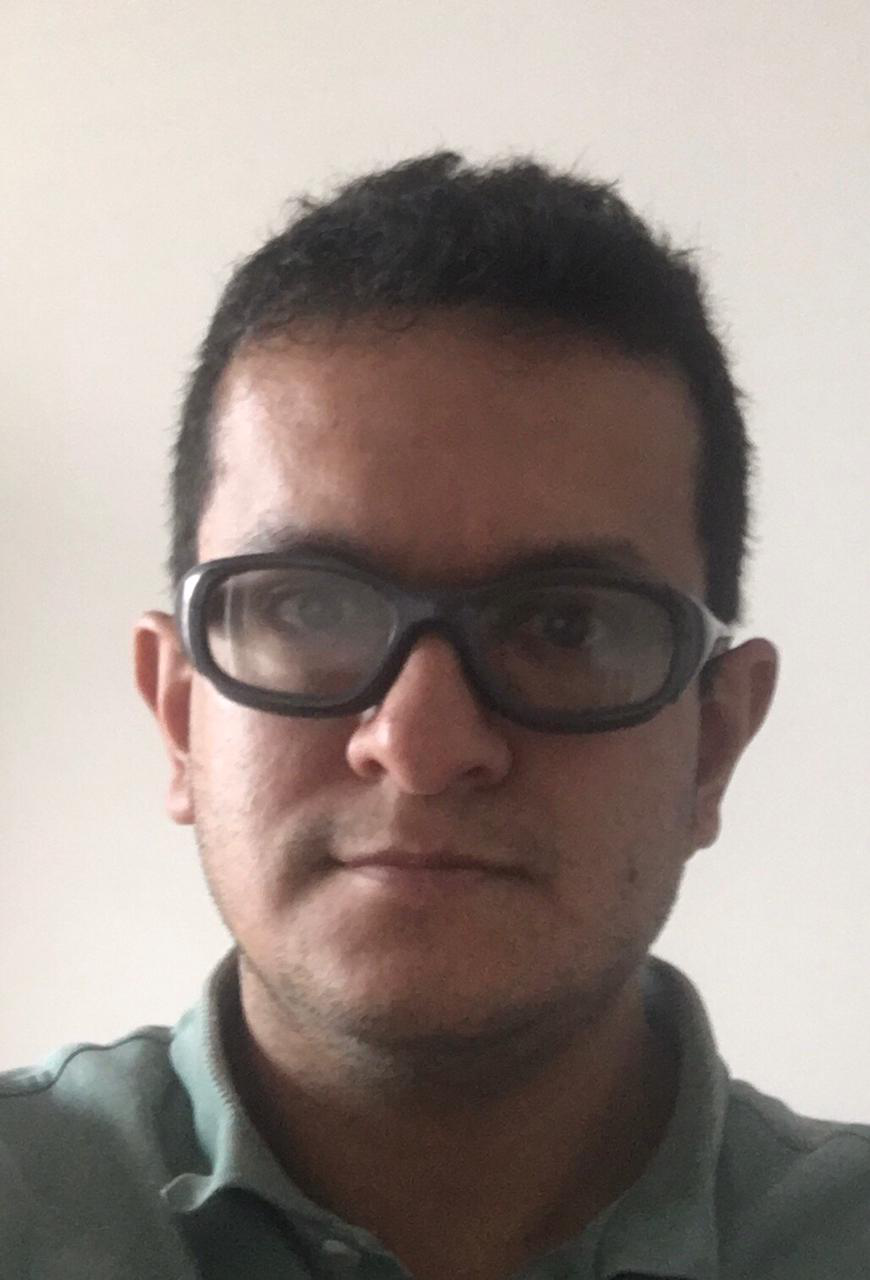}}]{Miguel Melgarejo}
(S'00-M'01-SM'11) received the BSc degree (with honors) in electronics engineering from the Universidad Distrital, Bogot\'a, Colombia, in 2001, the MSc  degree (with honors) in electronics engineering and computers from the De los Andes University, Bogot\'a, Colombia, in 2004 and the PhD degree (Magna Cum Laude) in engineering from the pontifical Xaverian University, Bogot\'a, Colombia in 2018. 
From 2001, he has been with the department of electronics engineering at Universidad Distrital and the laboratory for automation and computational intelligence, Bogotá, Colombia, where he is currently an associate professor. He has authored or coauthored over 80 technical papers. Prof. Melgarejo served as member of the technical program committee of the 2008 IEEE international conference on fuzzy systems and the technical program committee of the 2011 IEEE symposium on advances in type-2 fuzzy systems. His current research focuses on computational intelligence and complex systems.
\end{IEEEbiography}

\vspace{-.5cm}
\begin{IEEEbiography}[{\includegraphics[width=1in,height=1.25in,clip,keepaspectratio]{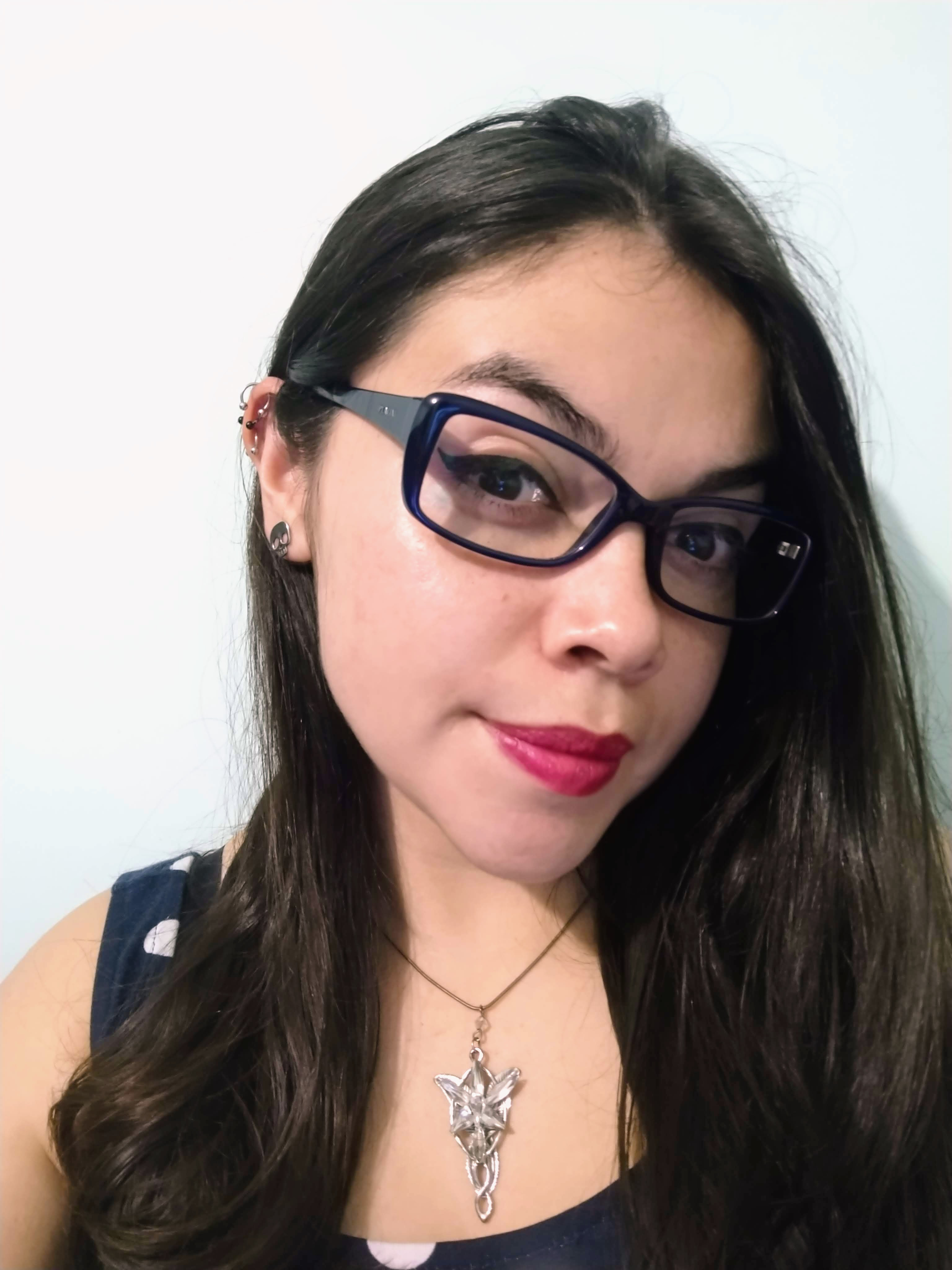}}]{Angie Blanco-Ca\~non}
Received the BSc degree in industrial engineering from the Universidad Distrital, Bogot\'a, Colombia, in 2016. Since 2017 is a student of the Master in Industrial Engineering in the Universidad Distrital. Her current research interests are on bio-inspired optimization algorithms applied to stochastic problems, hybrid algorithms, operations management scheduling models and machine learning.
\end{IEEEbiography}

\vspace{-.5cm}
\begin{IEEEbiography}[{\includegraphics[width=1in,height=1.25in,clip,keepaspectratio]{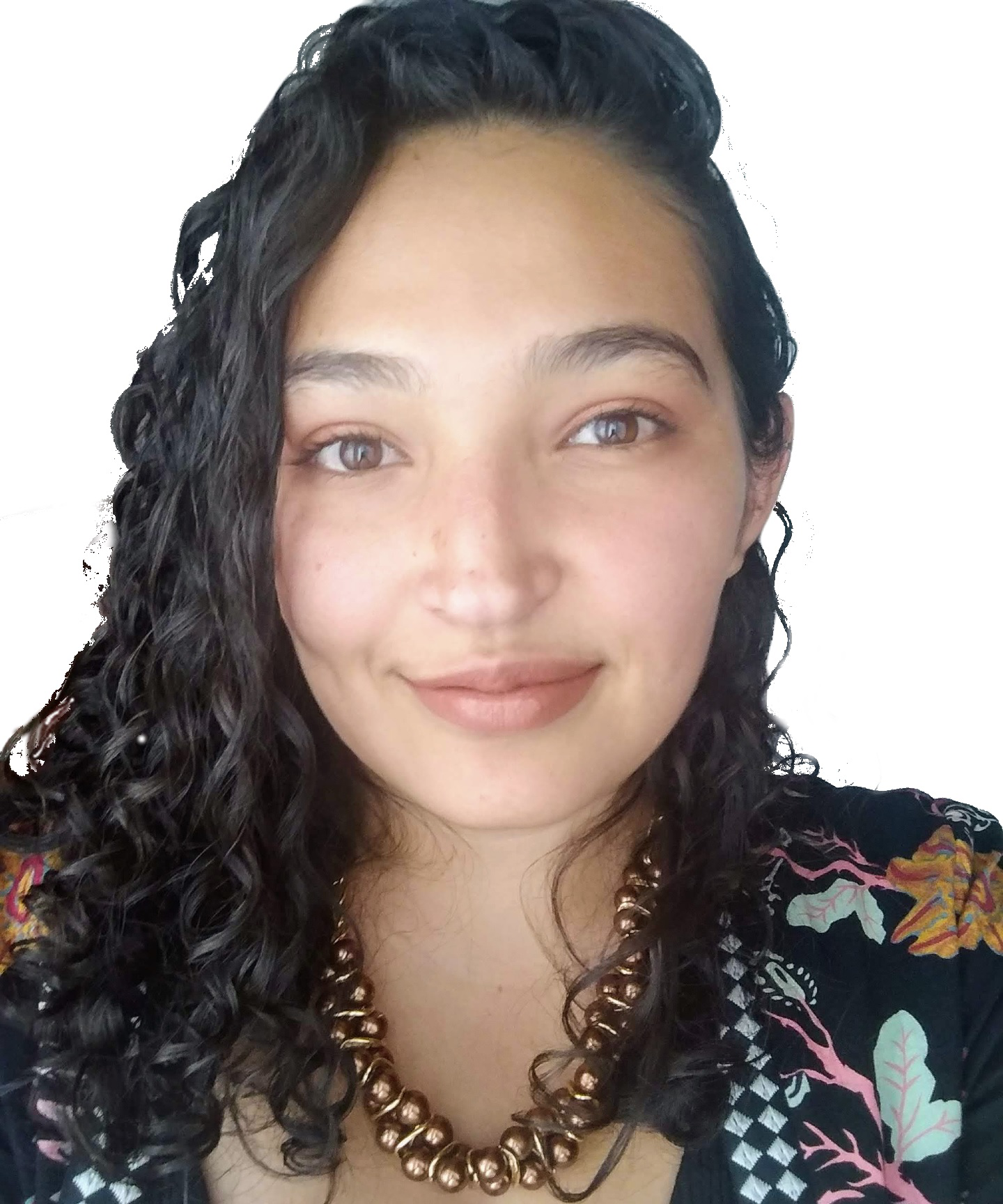}}]{Laura Reyes-Fajardo}
received the BSc degree in electronics engineering from the Universidad Distrital Francisco José de Caldas, Bogot\'a, Colombia, in 2016. Since 2017 is a student of the Master in Engineering in the Universidad Distrital. Her current research interests are on artificial intelligence applied to control systems, Boosting techniques and machine learning.
\end{IEEEbiography}

\vspace{-.5cm}
\begin{IEEEbiography}[{\includegraphics[width=1in,height=1.25in,clip,keepaspectratio]{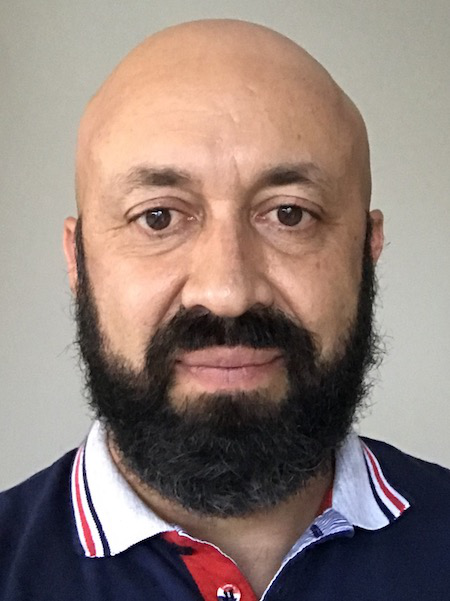}}]{Sergio Rojas-Galeano}
 received his Bachelor of Science (with honors) in Systems Engineering from Universidad Nacional de Colombia in 1998. He completed his Master of Science (with distinction) in Intelligent Systems from University College London (UCL) in 2005, and his PhD in Computer Science also from UCL in 2009. He is currently Full Professor at the School of Engineering of Universidad Distrital Francisco José de Caldas, Bogotá, Colombia. He held the position of Editor-in-Chief of \textit{Ingeniería} scientific journal during the period 2010-2018. His research interests include machine learning, bio-inspired computing, population-based metaheuristics, agent-based simulation and bibliometrics in engineering. He has authored or coauthored over 60 technical papers in these areas and about 20 editor's notes.
\end{IEEEbiography}






\end{document}